\documentclass[conference]{IEEEtran}
\usepackage{cite}
\usepackage{amsmath,amssymb,amsfonts}
\usepackage{algorithmic}
\usepackage{graphicx}
\usepackage{textcomp}
\usepackage{xcolor}

\usepackage[ruled,linesnumbered]{algorithm2e}
\usepackage{makecell}
\usepackage{float}
\usepackage{booktabs}
\usepackage{subfigure}
\usepackage{multirow}
\usepackage{graphbox}
\usepackage{cuted}
\usepackage{caption}
\usepackage[utf8]{inputenc}
\usepackage{enumitem}

\def\BibTeX{{\rm B\kern-.05em{\sc i\kern-.025em b}\kern-.08em
    T\kern-.1667em\lower.7ex\hbox{E}\kern-.125emX}}
\begin{document}

\title{MergeNet: Explicit Mesh Reconstruction from Sparse Point Clouds via Edge Prediction}
\author{Weimin Wang$^{1}$ \qquad Yingxu Deng$^{2}$ \qquad Zezeng Li$^{2}$ \qquad Yu Liu$^{1}$ \qquad Na Lei$^{1 *}$\\$^{1}$ DUT-RU International School of Information Science \& Engineering, Dalian University of Technology,  China
\\$^{2}$ School of Software, Dalian University of Technology,  China}

\maketitle
\footnotetext[1]{Corresponding author: Na~Lei (Email: nalei@dlut.edu.cn).} 

\begin{abstract}
This paper introduces a novel method for reconstructing meshes from sparse point clouds by predicting edge connection. 
Existing implicit methods usually produce superior smooth and watertight meshes due to the isosurface extraction algorithms~(e.g., Marching Cubes).
However, these methods become memory and computationally intensive with increasing resolution. 
Explicit methods are more efficient by directly forming the face from points. Nevertheless, the challenge of selecting appropriate faces from enormous candidates often leads to undesirable faces and holes.
Moreover, the reconstruction performance of both approaches tends to degrade when the point cloud gets sparse.
To this end, we propose MEsh Reconstruction
via edGE~(MergeNet), which converts mesh reconstruction into local connectivity prediction problems. 
Specifically, MergeNet learns to extract the features of candidate edges and regress their distances to the underlying surface. 
Consequently, the predicted distance is utilized to filter out edges that lay on surfaces. Finally, the meshes are reconstructed by refining the triangulations formed by these edges.
Extensive experiments on synthetic and real-scanned datasets demonstrate the superiority of MergeNet to SoTA
explicit methods.
\end{abstract}
\begin{IEEEkeywords}
Mesh reconstruction, Sparse point cloud, Connectivity prediction, Edge embedding
\end{IEEEkeywords}
\section{Introduction}
\label{sec:intro}

The increased accessibility of 3D sensors has significantly simplified the acquisition of point cloud data, which is now extensively utilized in environmental perception 3D reconstruction, and scene rendering. 
However, for subsequent applications such as robot manipulation, collision detection in robotics, or visualization and interaction in AR/VR, meshes are more essential. 
Therefore, reconstructing high-quality 3D meshes from point clouds has been a longstanding research topic aiming at cost-effective and efficient mesh generation.
Numerous endeavors have been advanced to tackle this task.
\begin{figure}[t]
  \centering
  \includegraphics[width=1\linewidth]{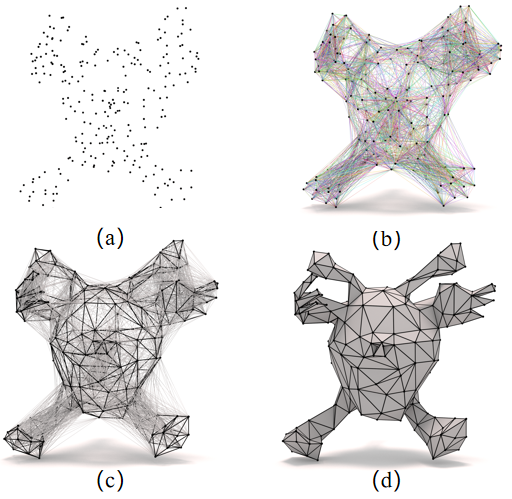}
  \vspace{-.3cm}
  \caption{The illustration of the pipelines of MergeNet. (a): sparse input point cloud; (b): candidate edges (colors indicate different edges); (c): edges selected by MergeNet~(thick lines); (d): refined mesh formed by filtered edges.
  }
  \vspace{-.7cm}
  \label{fig:banner}
\end{figure}

Traditional methods usually reconstruct the mesh either by explicitly connecting the points based on geometric rules~\cite{bernardini1999ball,edelsbrunner1994three} or extracting the surface from the formulated implicit representations~\cite{lorensen1987marching,carr2001reconstruction, kazhdan2006poisson,kazhdan2013screened}. 
Notably, the explicit Ball-pivoting algorithm (BPA)~\cite{digne2014analysis} and implicit Poisson surface reconstruction ~(PSR)~\cite{kazhdan2006poisson} have been widely utilized in the field.
Despite their practicality, these methods often require manual parameter tuning to accommodate various shapes.

Recently, data-driven learning-based methods have made notable success in 3D point cloud perception and understanding tasks.
Learning-based methods for mesh reconstruction are also proposed.
They design neural networks to learn either implicit representations such as Signed Distance Functions~(SDF)~\cite{petrov2023anise, chen2023unsupervised, wang2022rangeudf} or explicit geometric nature such as characteristics of triangles~\cite{sharp2020pointtrinet, liu2020meshing}.
Although learning-based methods demonstrate better performance than traditional ones, they are often limited by substantial computational demands and the performance would significantly degrade for sparse point clouds acquired by low-power 3D sensors in real scenarios.

To address these challenges, we propose a novel explicit approach for MEsh Reconstruction via edGE prediction~(MergeNet) to increase the deficiency advantages while improving the reconstruction performance.
In contrast to implicit representation-based approaches~\cite{petrov2023anise, wang2021neural, chen2023unsupervised, wang2022rangeudf}, the utilization of predicted edges directly in mesh reconstruction enhances efficiency rather than iso-surface extraction process.
Compared to existing explicit geometry-based methods~\cite{sharp2020pointtrinet, liu2020meshing}, the edge prediction design of MergeNet leads to a reduced candidate search space than face prediction, thereby facilitating the efficiency and accuracy of mesh generation. Contributions of this work are summarized as follows:
\begin{figure*}[ht!]
\vspace{-.4cm}
\centering
\includegraphics[width=.9\linewidth]{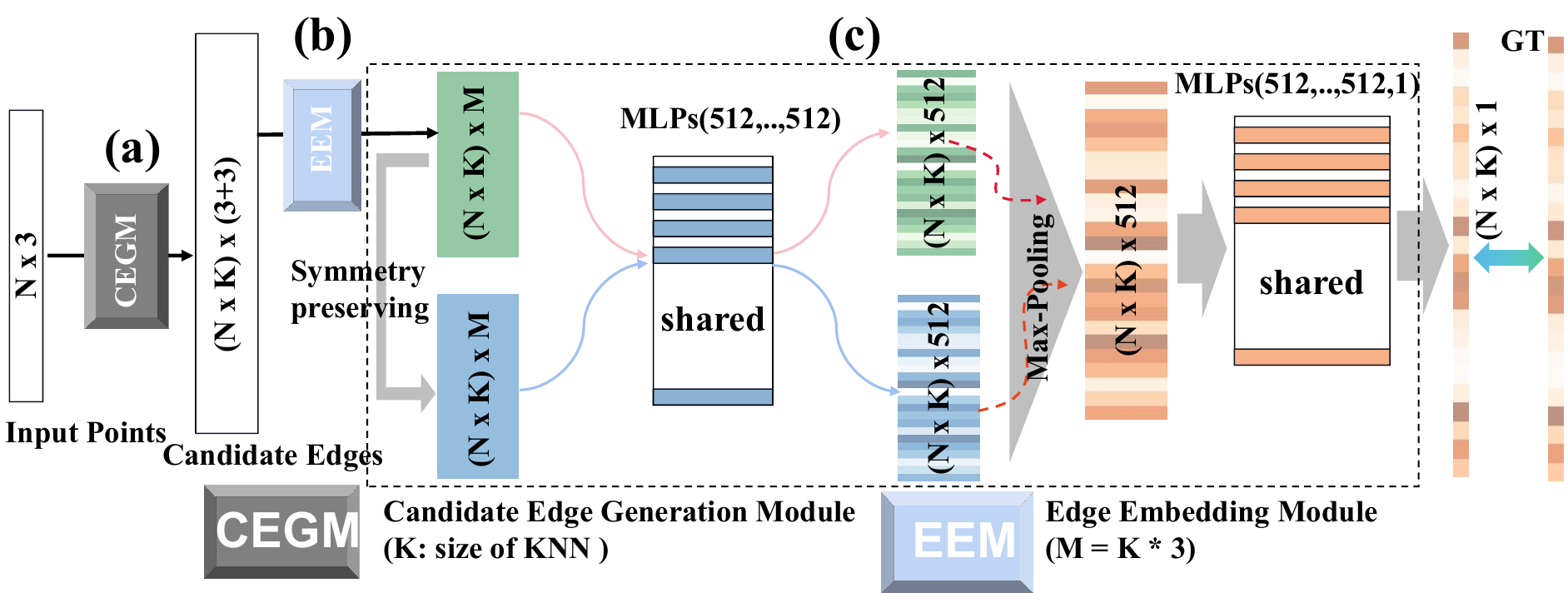}
\caption{Architecture of the proposed MergeNet which consists of three main steps: (a) Candidate edges generation module, (b) Edge embedding module and (c) Edge-to-surface distance regression loss. }
\label{fig:flowchart}
\end{figure*}

\vspace{-.2cm}
\begin{itemize}[leftmargin=*]
\setlength\itemsep{0em}
  \item We propose MergeNet, a novel edge-based approach for mesh reconstruction. It learns to predict edge connection directly, thereby enhancing efficiency and flexibility in mesh generation.
  \item To achieve the edge connectivity prediction, we introduce an edge canonical normalization strategy to embed the local features of edges and propose a line-to-surface loss to supervise the learning.

\item Extensive experiments on two datasets demonstrate the practicality and superiority of MergeNet. Remarkably, it outperforms all explicit baseline methods and even some learning-based implicit SoTA methods.
\end{itemize}

\section{RELATED WORK}
The utilization of 3D meshes in various fields results in the development of various reconstruction algorithms. In this section, we review them from the viewpoint of traditional geometric-based and data-driven learning-based methods.

\subsection{Geometric-based methods}
From the aspect of how meshes are generated,  reconstruction methods can be categorized into two streams: \textbf{explicit} and \textbf{implicit} reconstruction. Explicit approaches, such as Ball-pivoting algorithm (BPA)~\cite{bernardini1999ball}, Delaunay triangulation~\cite{boissonnat1993three} and Alpha shapes~\cite{edelsbrunner1994three}, and Zippering~\cite{turk1994zippered} rely on estimating local surface connectivity and connecting sampled points directly with triangles. 

Implicit reconstruction methods~\cite{kazhdan2006poisson,kazhdan2013screened,oztireli2009feature} aims to estimate the representation with a field function, such as a signed distance function. Consequently, iso-surfaces algorithm such as Marching Cubes is utilized to extract surfaces. 
However, solving large-scale equations is usually time-consuming.
Furthermore, the performance may greatly depend on the accuracy of other geometric information~(e.g., normals). 
\subsection{Learning-based methods}
Learning-based methods benefit from the continuous growth of open datasets and can be divided into two categories~\cite{lei2023whats}: explicit and implicit approaches. 
Implicit approaches~\cite{maruani2023voromesh, chen2023unsupervised, camps2022learning, BaoruiMa2023ReconstructingSF} utilize deep neural networks to learn the implicit representation and then extract isosurfaces using the Marching Cubes algorithm~\cite{lorensen1987marching}. As a result, these approaches encounter challenges such as high computational complexity.
The explicit approaches~\cite {liu2020meshing, sharp2020pointtrinet, hanocka2020point2mesh, zhang2023dmnet} leverage deep neural networks to directly learn to predict geometric primitives. 

For instance, IER~\cite{liu2020meshing} proposes a neural network to predict the ratio of intrinsic and extrinsic Euclidean distance between points, which is further used to guide the generation of the triangular mesh. PointTriNet~\cite{sharp2020pointtrinet} iteratively updates two networks: a classification network that predicts whether a candidate triangle is reasonable, and a proposal network that suggests additional candidate triangles.
Point2Mesh~\cite{hanocka2020point2mesh} takes the input point cloud as the self-prior and deforms an initial convex mesh by optimizing the weights of the networks to shrink-wrap to a desirable solution. 
Additionally, in a more recent development, the concept of diffusion is also utilized for 3D reconstruction~\cite{lyu2023controllable}.

However, learning-based methods typically demand substantial computation resources. On the other side, the performance of the models may drastically degrade when dealing with sparse point clouds. 
To address the challenges, we propose the learning-based explicit reconstruction approach that reformulates mesh reconstruction as an edge prediction problem. It learns to select candidate edges which makes it more efficient.
Moreover, owing to the normalization strategy of local features, our approach exhibits better generability and robustness to the sparse point cloud.

\section{Methodology}

Given a point cloud $\mathcal{P}$ with $N$ points, we aim to reconstruct a triangulated mesh $M = (V, F)$, where $V=\mathcal{P} \in \mathbb{R}^3$ represents the vertex set and $F$ denotes the set of faces. Our key insight is to regard surface reconstruction as an edge-face-distance regression task. Edges $E$ are predicted as connections between a vertex and its neighborhoods.
As illustrated in Fig~\ref{fig:flowchart}, the proposed method mainly consists of three steps that are depicted from \S~3.1 to 3.3.

\subsection{Candidate edge generation}
For a point cloud $\mathcal{P}$, MergeNet learns to select the right edge from the candidate space of maximum $C_N^2$ edges.
The candidate edge generation module~(\textbf{CEGM}) needs to take the trade-off between coverage and efficiency, that is,
the number of edges should be large enough to cover all ground truth edges and small enough to keep efficient.
With the observation of the correlation between connectivity and distance, CEGM produces $N\cdot K$ edges for each vertice in this work.

\subsection{Edge embedding by local canonical normalization}
\textbf{Transformation-invariance.} 
Since MergetNet aims to learn to predict the validity of candidate edges, the embedding of each edge together with its local geometric information is essential for training the network. 
A straightforward way is to concatenate the coordinates of endpoints and their $n$ nearest neighboring points. 
However, the generalization ability would be significantly limited.
Edge embedding module~(\textbf{EEM}) needs to be transformation-invariant. Thus, we propose a canonical normalization approach by forming a normalized coordinate system based on the vector of the edge $e_j^i$ and the centroid of its $n$ nearest neighboring points $\mathcal{N}(e_j^i$). The distance from one point to the edge is defined by its distance to the midpoint of the edge.

\textbf{Canonical normalization}. 
To form the canonical normalized coordinate system, it is necessary to determine the origin, one axis, a plane and orientation of axes~(e.g., left-hand rule).
\begin{figure}[hbpt]
  \centering
  \includegraphics[width=1\linewidth]{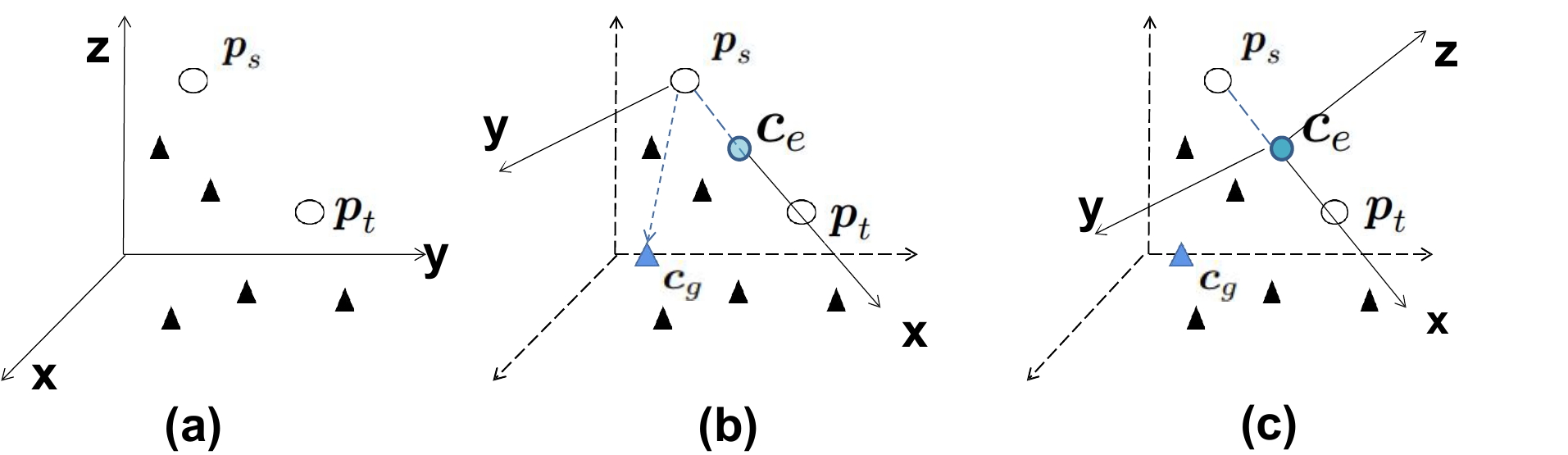}
    \caption{The illustration of local canonical normalization. (a) Example edge with endpoints $\boldsymbol{p}_s$ and $\boldsymbol{p}_t$. (b)  Midpoint $\boldsymbol{c}_e$ of the edge as origin and the plane of $\mathbf{\boldsymbol{p}_s\boldsymbol{p}_t\boldsymbol{c}_g}$ as the $XY$ plane.  (c) Canonical coordinate systems with left-hand rule.}
  \label{fig:trans}
\end{figure}

As illustrated in Fig.~\ref{fig:trans}, for an edge starting from $\boldsymbol{p}_s$ to $\boldsymbol{p}_t$, we define the origin as the midpoint  $\boldsymbol{c}_e = \frac{\boldsymbol{p}_s + \boldsymbol{p}_t}{2}$. The direction of  $\overrightarrow{\mathbf{\boldsymbol{c}_e \boldsymbol{p}_t}}$ is taken as $X$-axis. To decide a unique plane as $XY$ plane, we introduce the vector from $\boldsymbol{p}_s$ to the centroid of $n$ nearest neighboring points $\boldsymbol{c}_g$. Consequently, the $Y-$ and $Z-$axis can be determined by the left-hand rule as follows:

\begin{equation}
\mathbf{y} = \overrightarrow{\mathbf{\boldsymbol{p}_s \boldsymbol{p}_t}} \times \overrightarrow{\mathbf{\boldsymbol{p}_s \boldsymbol{c}_g}}
\end{equation}
\begin{equation}
\mathbf{z} = \mathbf{x} \times \mathbf{y}\ .
\label{eq:zaxis}
\end{equation}
where $\times$ represents the cross product.



Finally, neighbor points are transformed into the canonical system. The normalized coordinates are flattened into a $M$-D vector as the embedding of local geometric information. We denote the embedding with the canonical normalization as $E_{mb}({e}_j^i, \mathcal{N}(e_j^i)$.


\subsection{Network for regression of edge-to-surface distance }

\textbf{Symmetry preservation.} 
If the connection between $\boldsymbol{p}_i\boldsymbol{p}_j$ is a valid edge, $\boldsymbol{p}_j\boldsymbol{p}_i$ should also be the same. 
However, EEM would embed differently for the two directions.
To maintain the symmetry, we introduce the symmetric embedded feature of the symmetric directions. Specifically, we reverse the sign of the \textbf{x} and \textbf{y} coordinates of all local points due to the reverse of $X$-axis of the canonical coordinate system.


\textbf{Edge-to-surface distance.} 
With the embedded local geometric features, we first feed them to the MLPs  to extract high-level features, as illustrated in (c) part of Fig.~\ref{fig:flowchart}.
A max-pooling is then applied to the extracted features of original input and symmetric one for symmetry preserving.
The operation of feature extraction and maxpooling is denoted as $f_\theta$.
The features are fed to another MLPs $g_\phi$ for the edge-to-surface distance regression. 
In this work, we sample 10 points on the edge with an equal interval. The edge-to-surface distance is defined as the maximum point-to-surface distance.

\subsection{Loss functions}
We attempted to treat the point connectivity as a classification problem, taking the set of mesh edges of the manifold as the ground truth, and designed a neural network to predict the point connections. However, the network failed to optimize with the limited supervision. 
We transform the problem into one of predicting the edges close to the surface of the object. To achieve this, we design our neural network to learn the distance of edges to the surface of the object. We use the squared error to minimize the difference between the ground truth distance and the distance output by the network.
In this way, we can train the network to accurately predict the edges close to the surface of the mesh, which can then be used to reconstruct the surface via post-processing.



\subsection{Mesh generation from edges}

The network outputs the connection relationship between two points to provide more flexibility for subsequent processing.
Candidate edges with side lengths exceeding 1.5 times the average will be deleted. Three edges form a triangle if they form a closed loop.
Triangles are sorted in ascending order according to the length of edges. If the current triangle does not overlap with triangles in the existing set, it is added to the set. Edges appearing only once in triangles are identified, and rings are located based on these edges. The rings are converted into triangles and added to the collection of triangles.

\begin{table}[h]
\caption{Quantitative results on ShapeNet.}
\begin{tabular}{cc|cccc}
\hline
\multicolumn{2}{l|}{ } & \makecell[c]{$\mathrm{L_1 CD}\downarrow$ \\ ($\times 10^{-2}$)}     & \makecell[c]{$\mathrm{L_2 CD}\downarrow$ \\ ($\times 10^{-4}$)} & F-Score$\uparrow$ & NC$\uparrow$ \\ 
\hline
\multicolumn{1}{l|}{\multirow{2}{*}{\makecell[c]{Implicit }}} & \makecell[c]{PSR} & 12.35 & 227.86  & 0.41 & 0.65\\ 
\multicolumn{1}{l|}{}                                          & \makecell[c]{On-Surface}  & \textbf{0.68}  & - & \textbf{0.95}  & \textbf{0.89}       \\ \hline
\multicolumn{1}{l|}{\multirow{3}{*}{Explicit}} & \makecell[c]{BPA}        & 1.85  &  4.24 & 0.77 & 0.79   \\ 
\multicolumn{1}{l|}{}                                          & \makecell[c]{PointTriNet} & -   & 2.72 & 0.92 & 0.88 \\ 
\multicolumn{1}{l|}{}                                          & \makecell[c]{IER}    & -  & 8.63 & 0.87 & 0.68 \\ 
\multicolumn{1}{l|}{}                                          & \makecell[c]{Ours}  & \underline{1.26} & \underline{1.14} & \underline{0.93} & \textbf{0.89} \\ \hline
\end{tabular}
\label{L1_ShapeNet}
\end{table}

\begin{table}[ht]
\caption{Quantitative results on SHREC.}
\resizebox{\linewidth}{!}{
\begin{tabular}{cc|cccc}
\hline
\multicolumn{2}{l|}{}  & \makecell[c]{$\mathrm{L_1 CD}\downarrow$ \\ ($\times 10^{-2}$)}  & \makecell[c]{$\mathrm{L_2 CD}\downarrow$ \\ ($\times 10^{-4}$)} & F-Score$\uparrow$ & NC$\uparrow$ \\ 
\hline
\multicolumn{1}{l|}{\multirow{2}{*}{\makecell[c]{Implicit 
}}} & \makecell[c]{PSR} & 26.14 & 894.37 & 0.31 & 0.50  \\ 
\multicolumn{1}{l|}{}                                          & \makecell[c]{On-Surface}  & 4.48 &  19.83 & 0.89 & 0.84 \\ 
\hline
\multicolumn{1}{l|}{\multirow{3}{*}{\makecell[c]{Explicit
}}} & \makecell[c]{BPA}         & 2.74  & 9.33 & 0.33 & 0.81   \\
\multicolumn{1}{l|}{}                                          & \makecell[c]{PointTriNet} & \underline{2.42} & \underline{6.86} & \underline{0.92} & \textbf{0.87}   \\ 
\multicolumn{1}{l|}{}                                          & \makecell[c]{IER}         & 4.03 & 15.28 & 0.87 & 0.73  \\ 
\multicolumn{1}{l|}{}                                          & Ours            & \textbf{1.98} & \textbf{3.03} & \textbf{0.93} & \underline{0.85}   \\
\hline
\end{tabular}
}
\label{L2_SHREC}
\end{table}


\begin{figure*}[ht]
\centering
\begin{minipage}{0.115\linewidth}
\centerline{\includegraphics[width=1.2\textwidth]{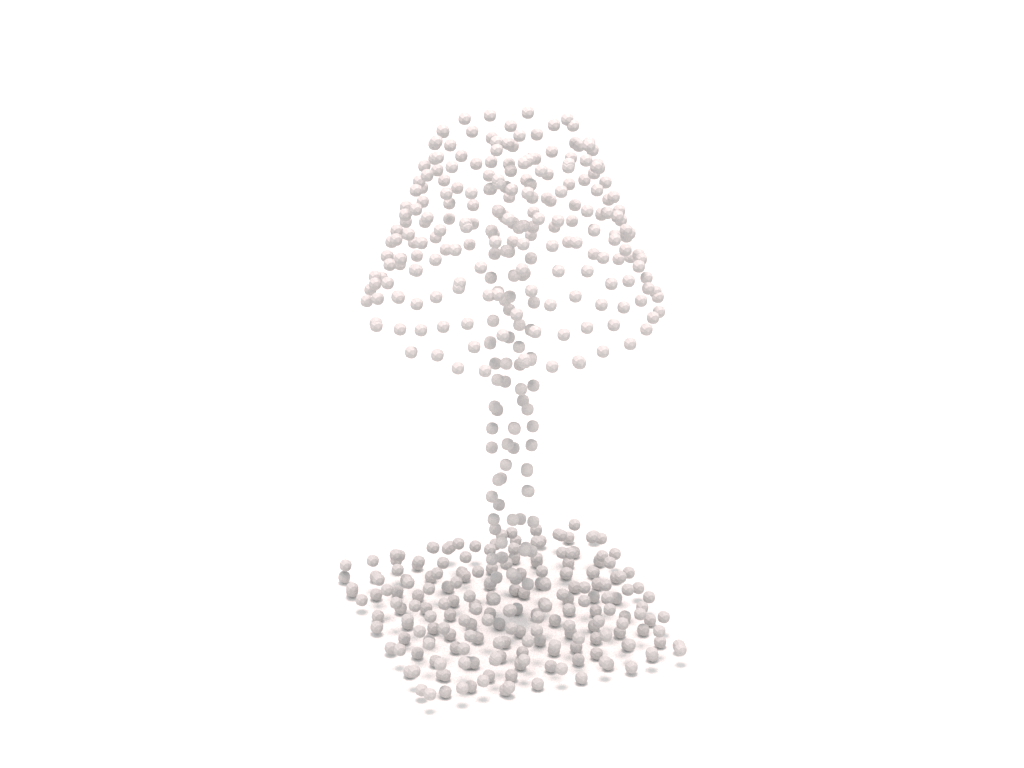}}
\vspace{-.1cm}
\centerline{\includegraphics[width=1.2\textwidth]{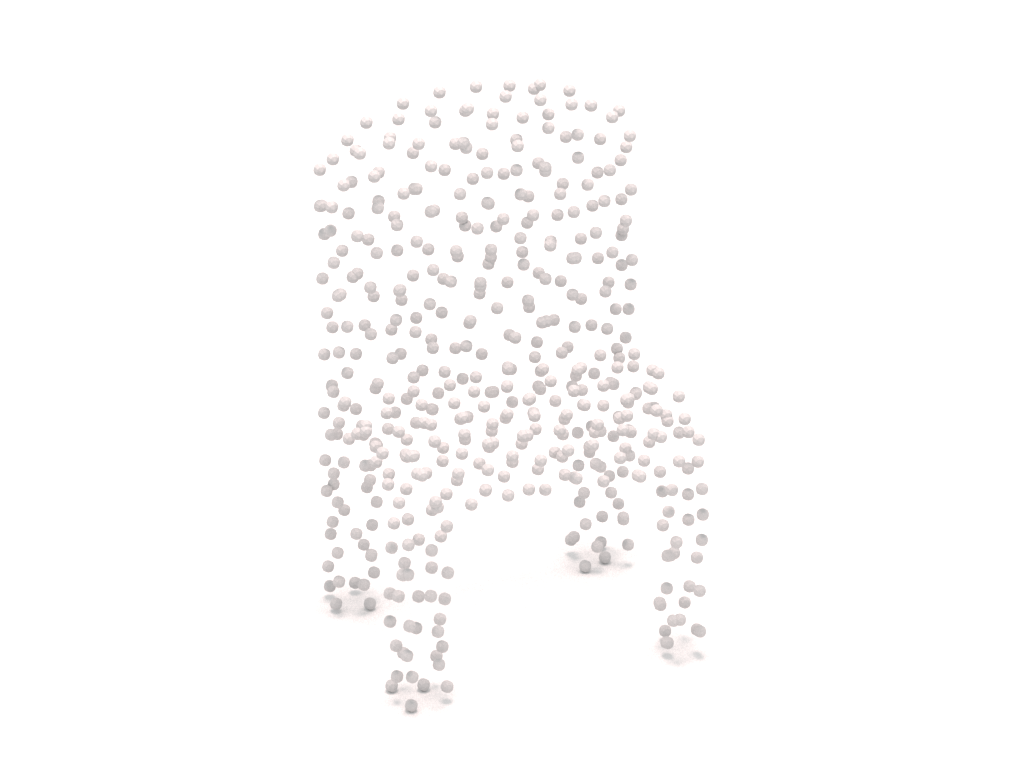}}
\vspace{-.1cm}
\centerline{\includegraphics[width=1.2\textwidth]{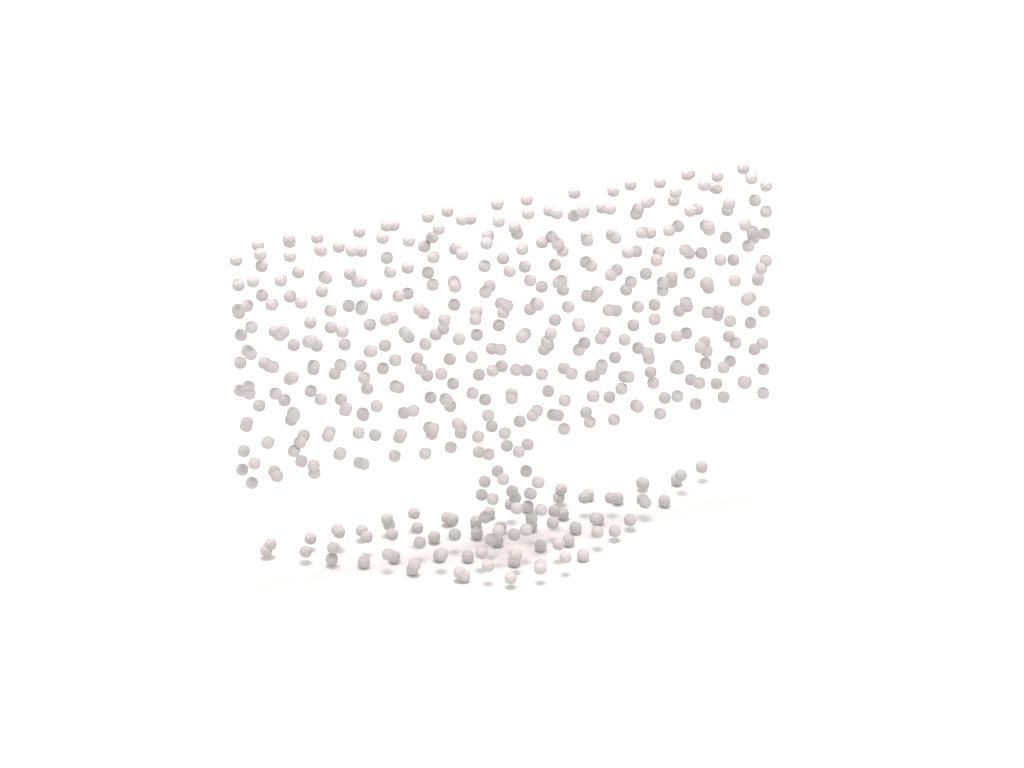}}
\vspace{-.1cm}
\vspace{-.1cm}
\centerline{\includegraphics[width=1.2\textwidth]{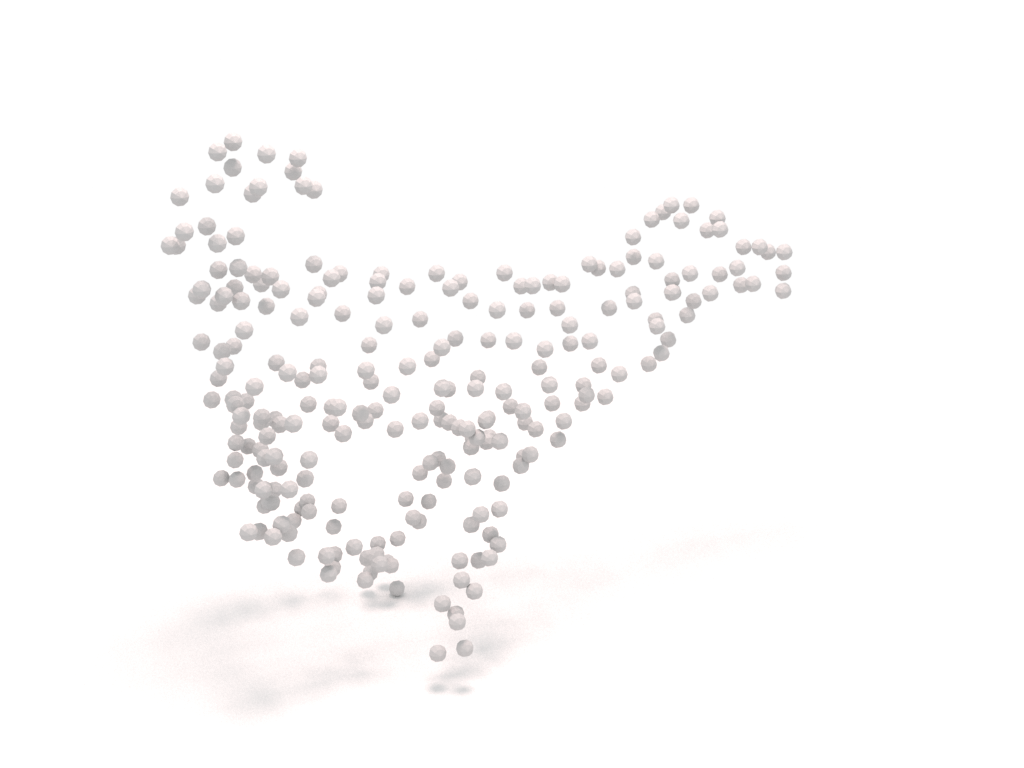}}
\centerline{\scriptsize {\makecell[c]{Input \\ \\}}}
\end{minipage}
\begin{minipage}{0.115\linewidth}
\centerline{\includegraphics[width=1.2\textwidth]{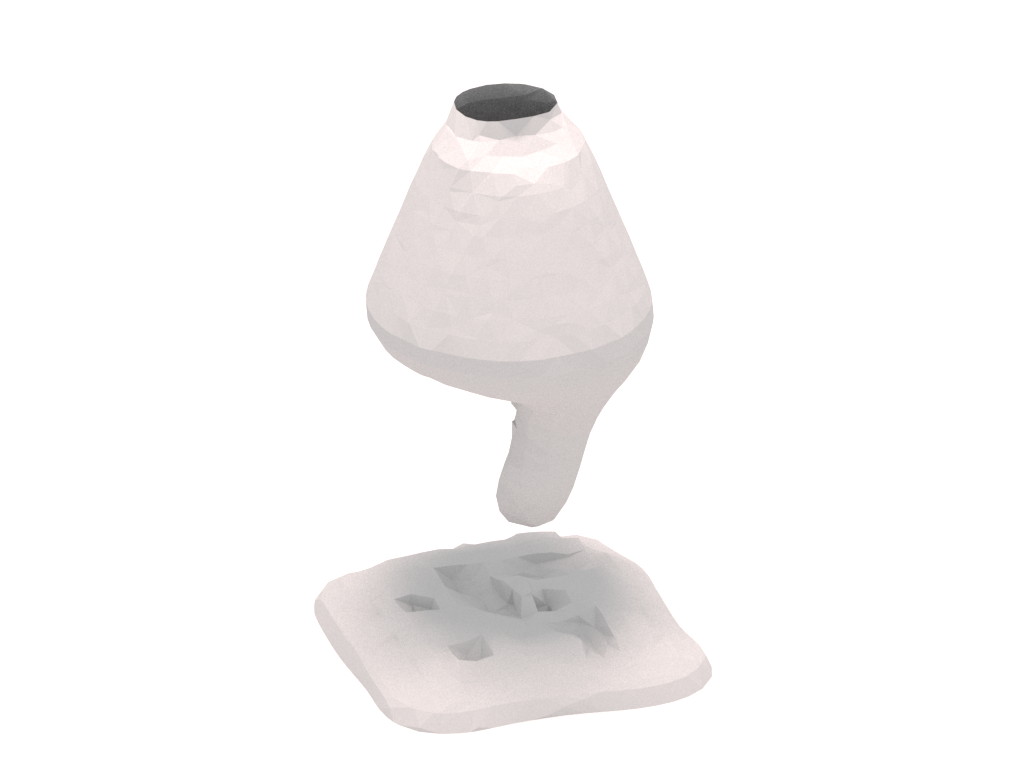}}
\vspace{-.1cm}
\centerline{\includegraphics[width=1.2\textwidth]{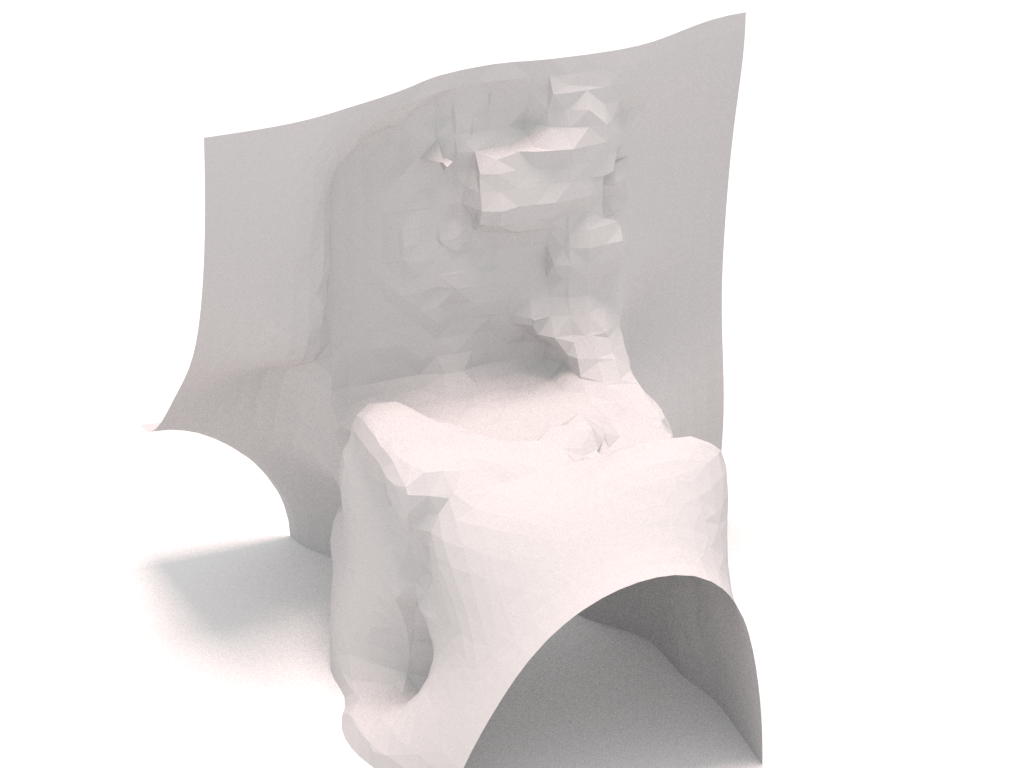}}
\vspace{-.1cm}
\centerline{\includegraphics[width=1.2\textwidth]{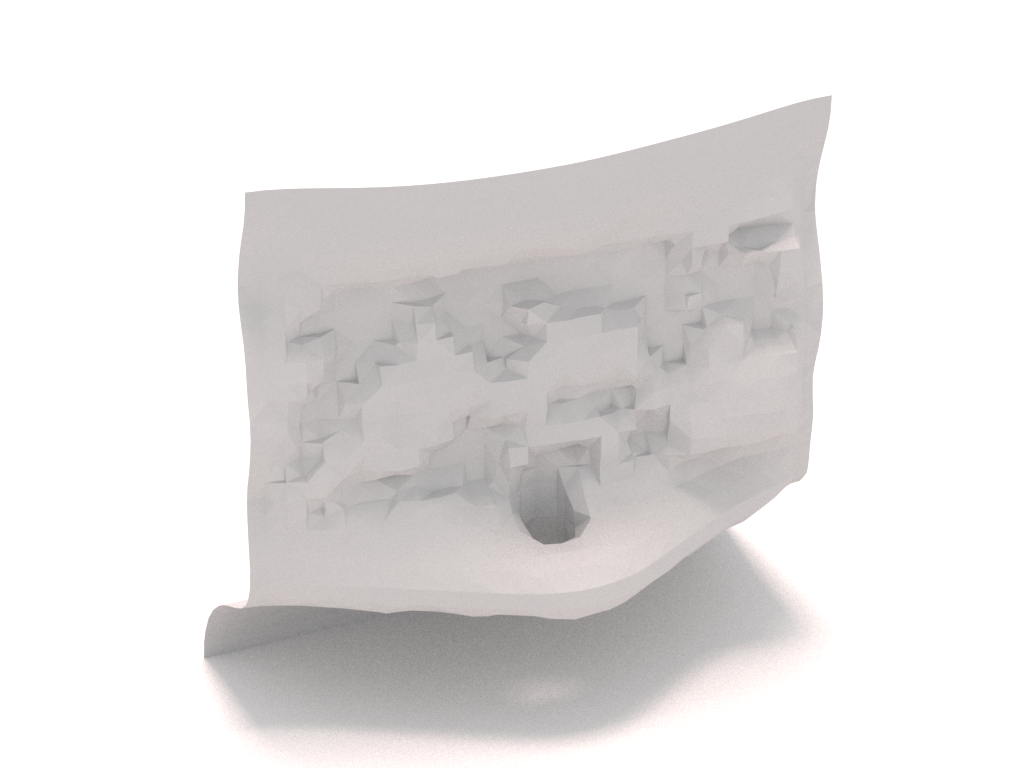}}
\vspace{-.1cm}
\vspace{-.1cm}
\centerline{\includegraphics[width=1.2\textwidth]{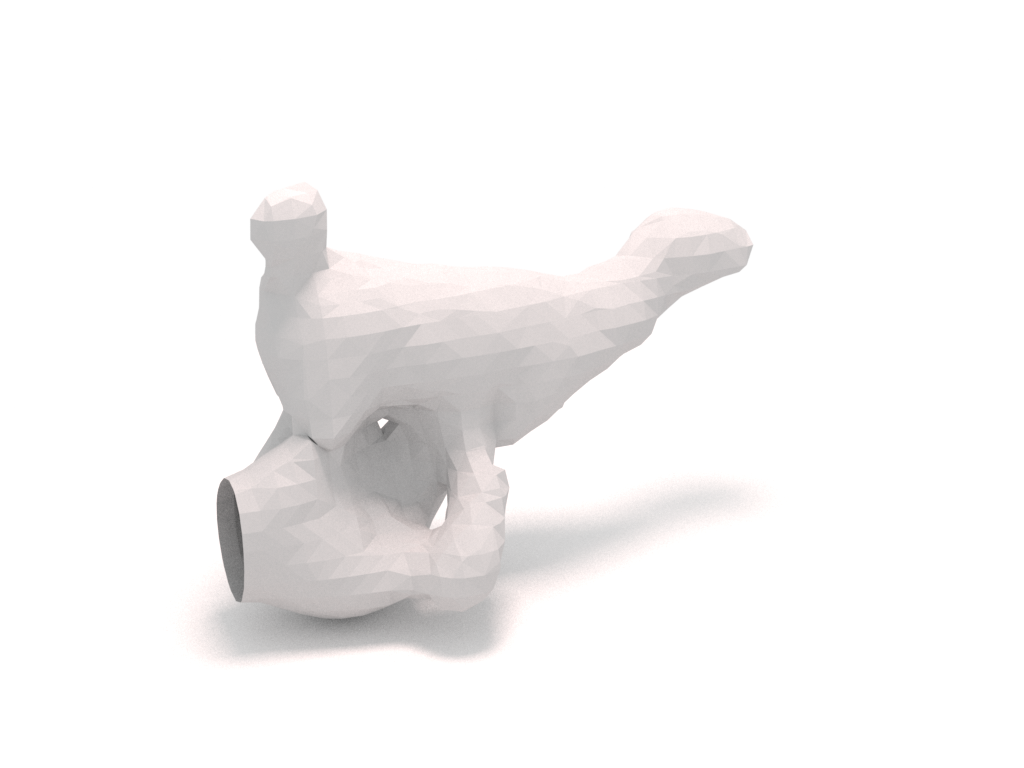}}
\centerline{\scriptsize{\makecell[c]{PSR \\ \\}}}
\end{minipage}
\begin{minipage}{0.115\linewidth}
\vspace{-.2cm}
\centerline{\includegraphics[width=1.2\textwidth]{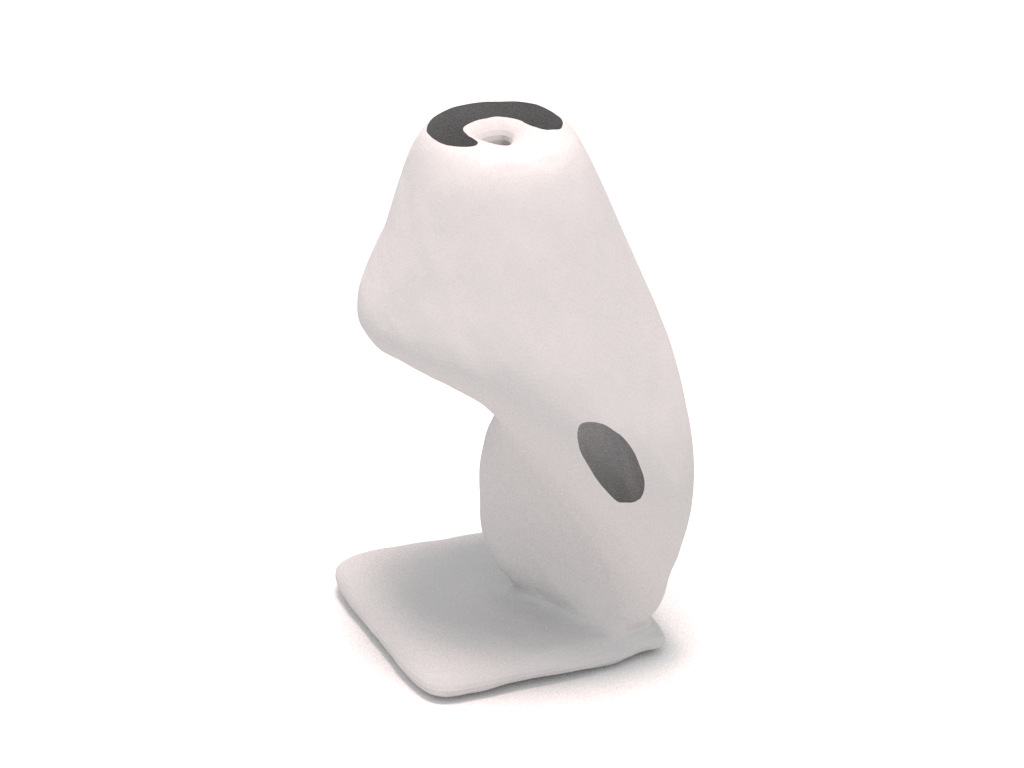}}
\vspace{-.2cm}
\centerline{\includegraphics[width=1.2\textwidth]{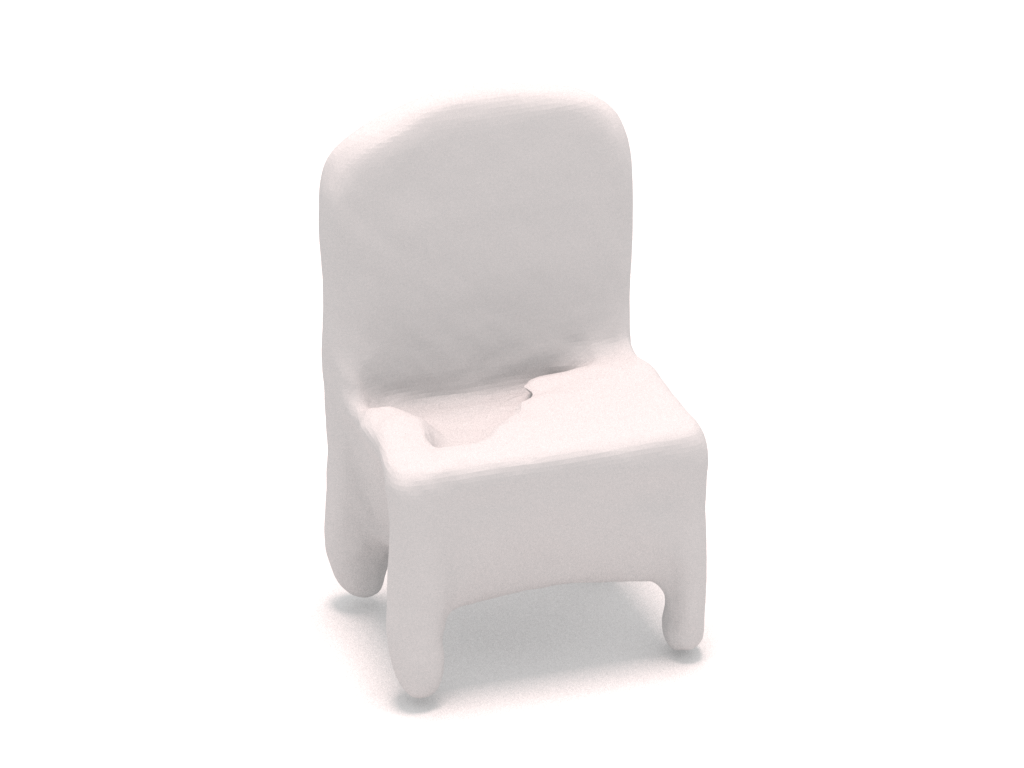}}
\vspace{-.2cm}
\centerline{\includegraphics[width=1.2\textwidth]{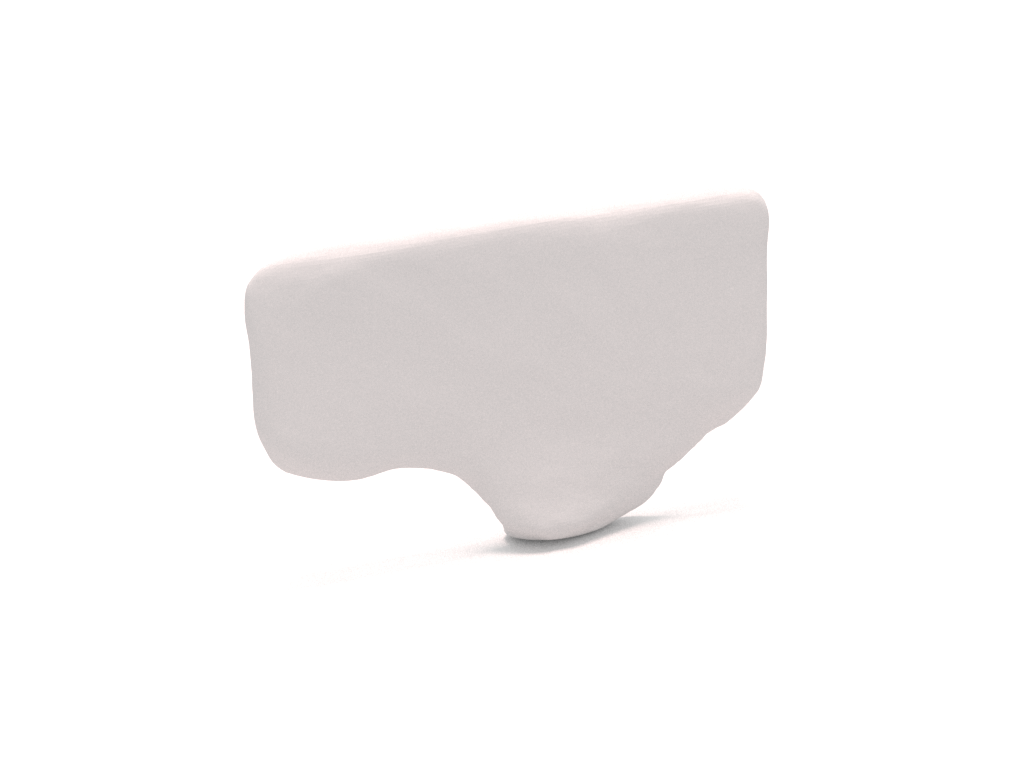}}
\vspace{-.1cm}
\vspace{-.1cm}
\centerline{\includegraphics[width=1.2\textwidth]{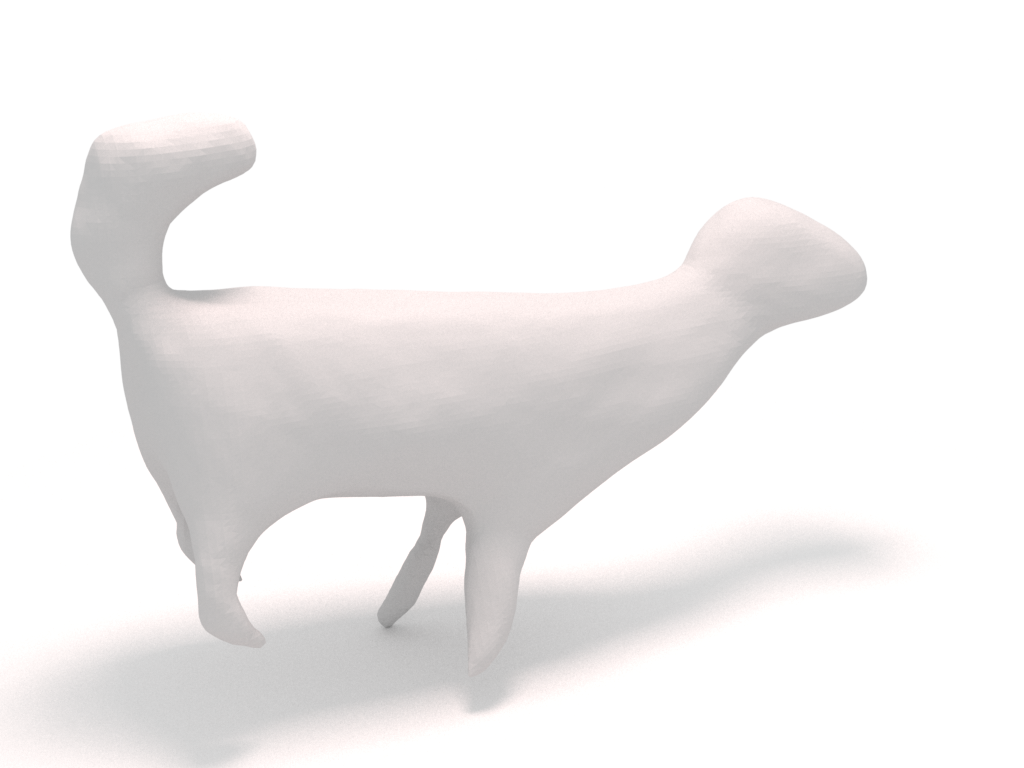}}
\centerline{\scriptsize{\makecell[c]{On-Surface}}}
\end{minipage}
\begin{minipage}{0.115\linewidth}
\centerline{\includegraphics[width=1.2\textwidth]{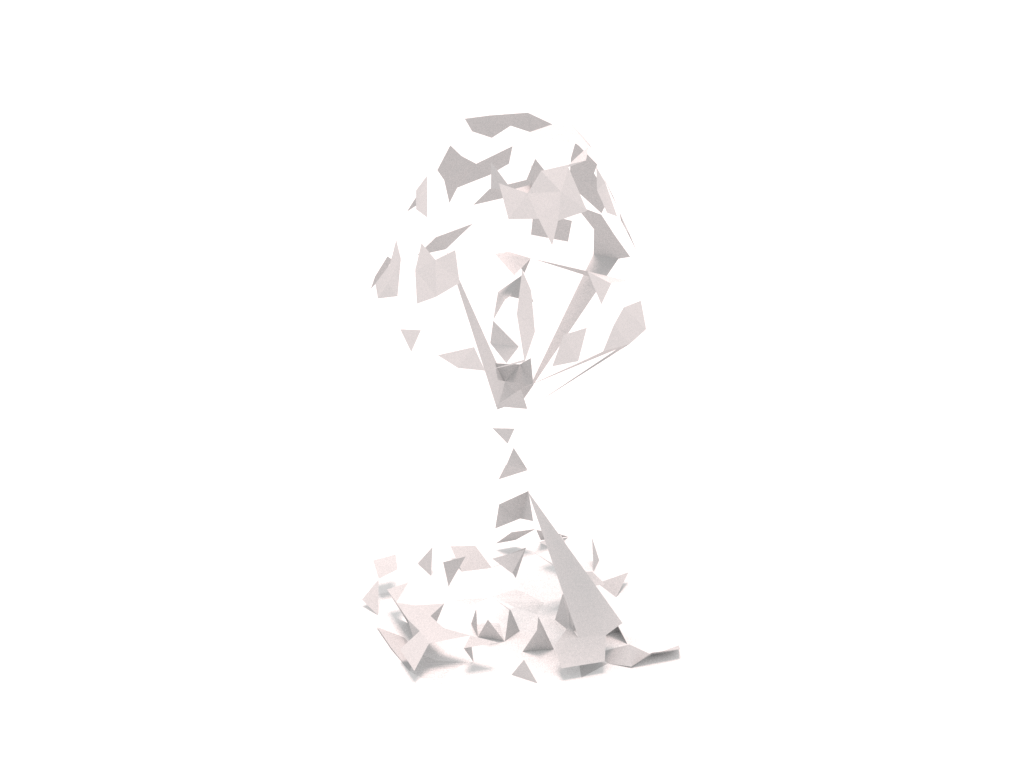}}
\vspace{-.1cm}
\centerline{\includegraphics[width=1.2\textwidth]{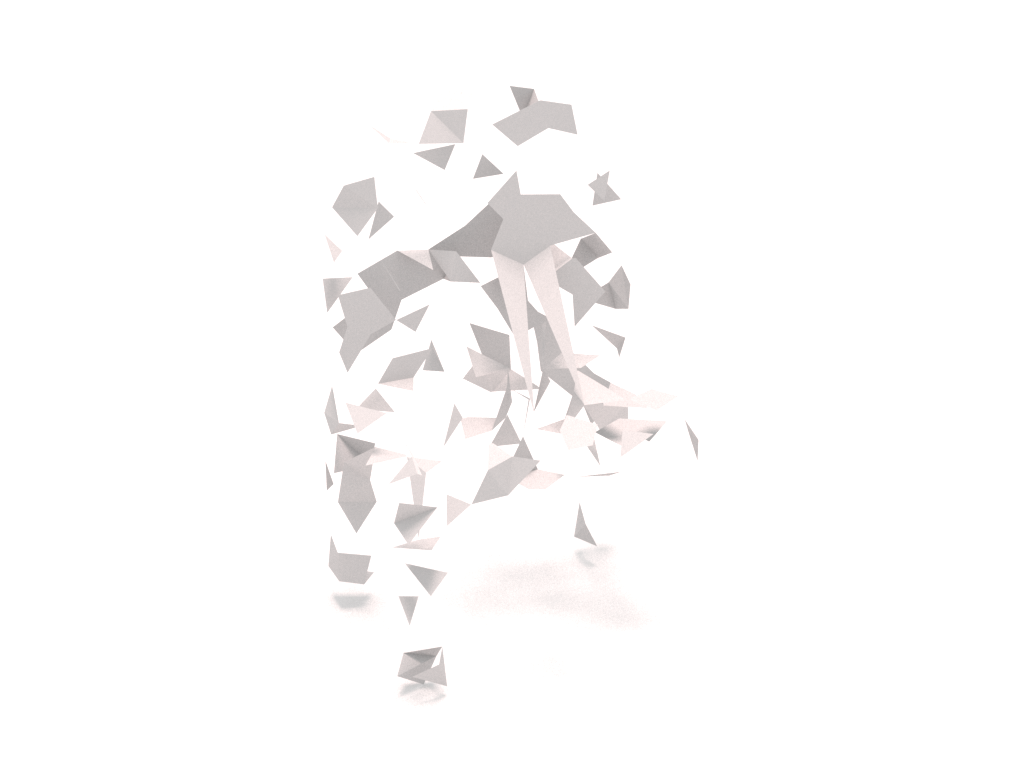}}
\vspace{-.1cm}
\centerline{\includegraphics[width=1.2\textwidth]{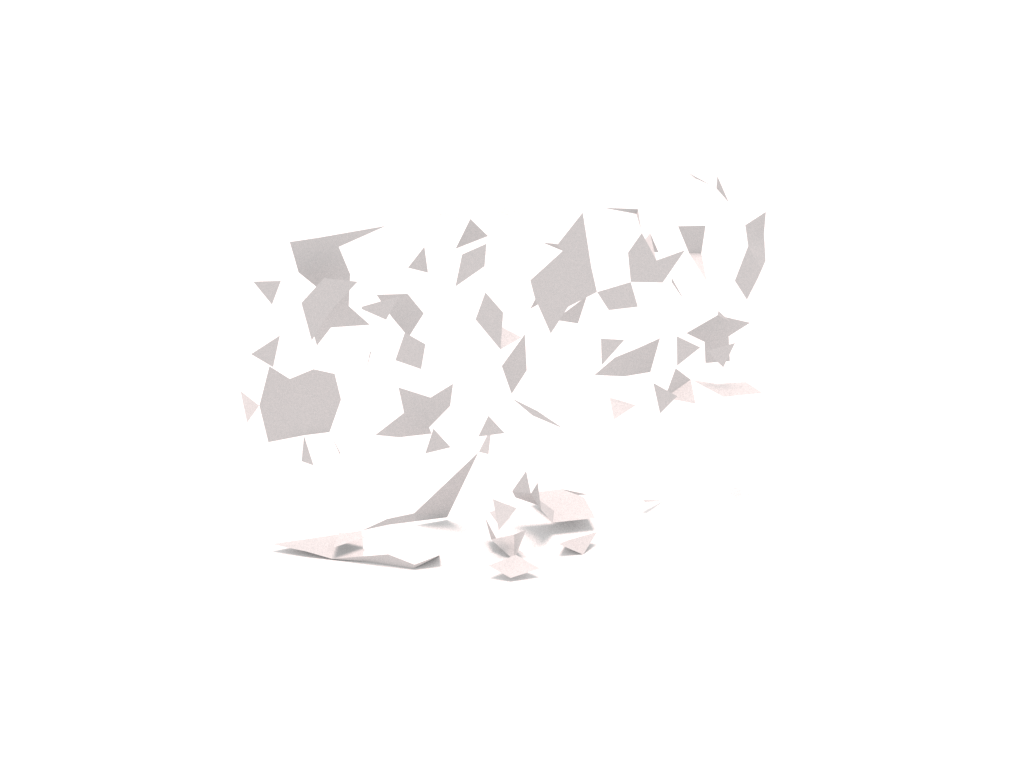}}
\vspace{-.1cm}
\vspace{-.1cm}
\centerline{\includegraphics[width=1.2\textwidth]{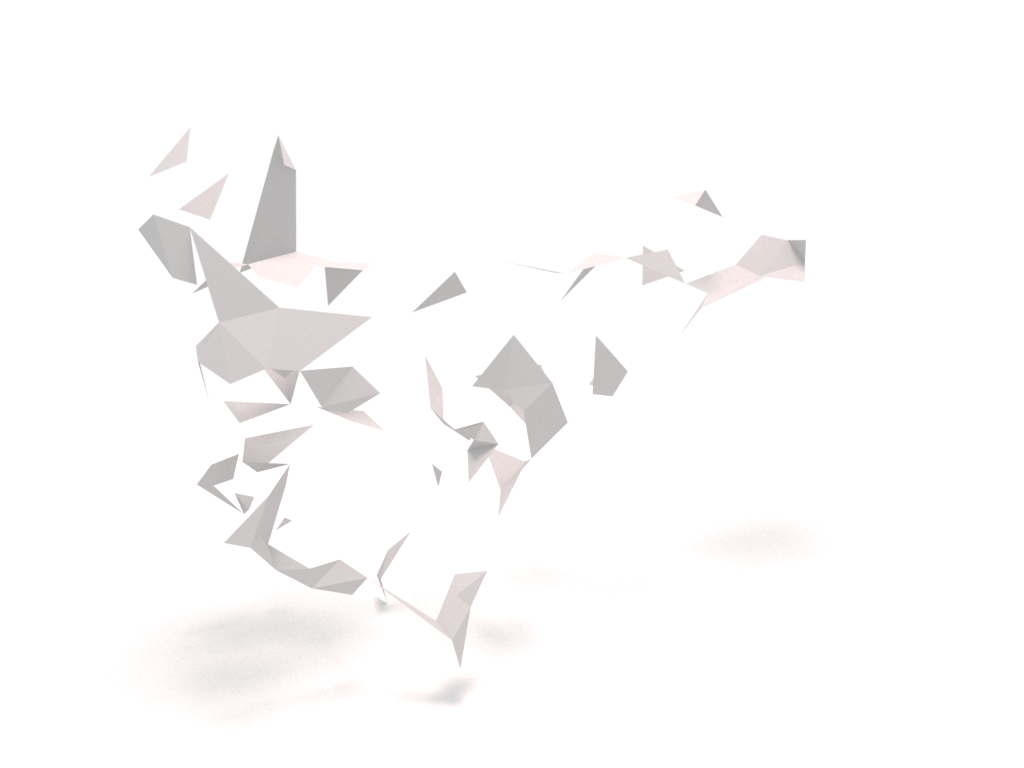}}
\centerline{\scriptsize {\makecell[c]{BPA \\ \\}}}
\end{minipage}
\begin{minipage}{0.115\linewidth}
\centerline{\includegraphics[width=1.2\textwidth]{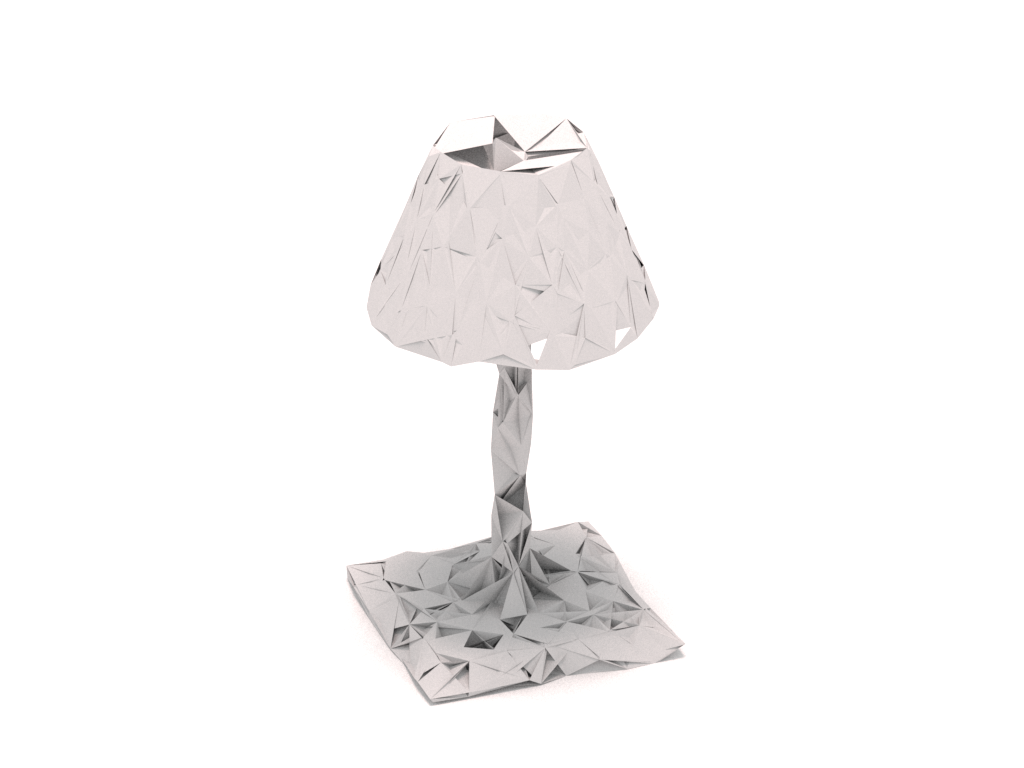}}
\vspace{-.1cm}
\centerline{\includegraphics[width=1.2\textwidth]{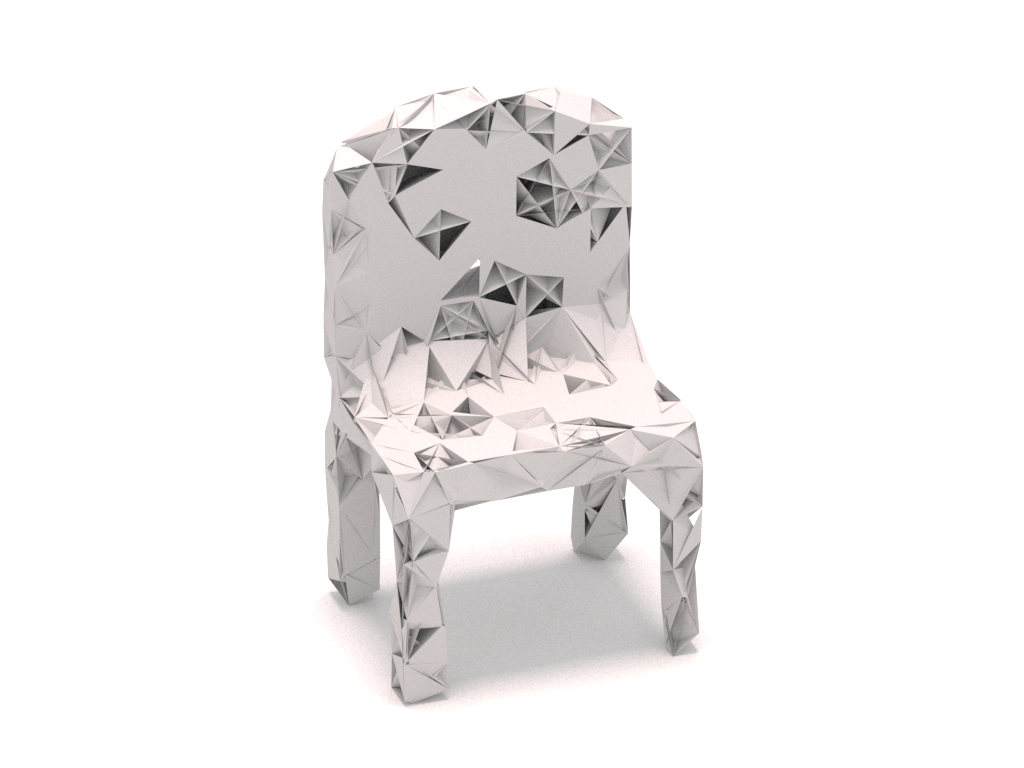}}
\vspace{-.1cm}
\centerline{\includegraphics[width=1.2\textwidth]{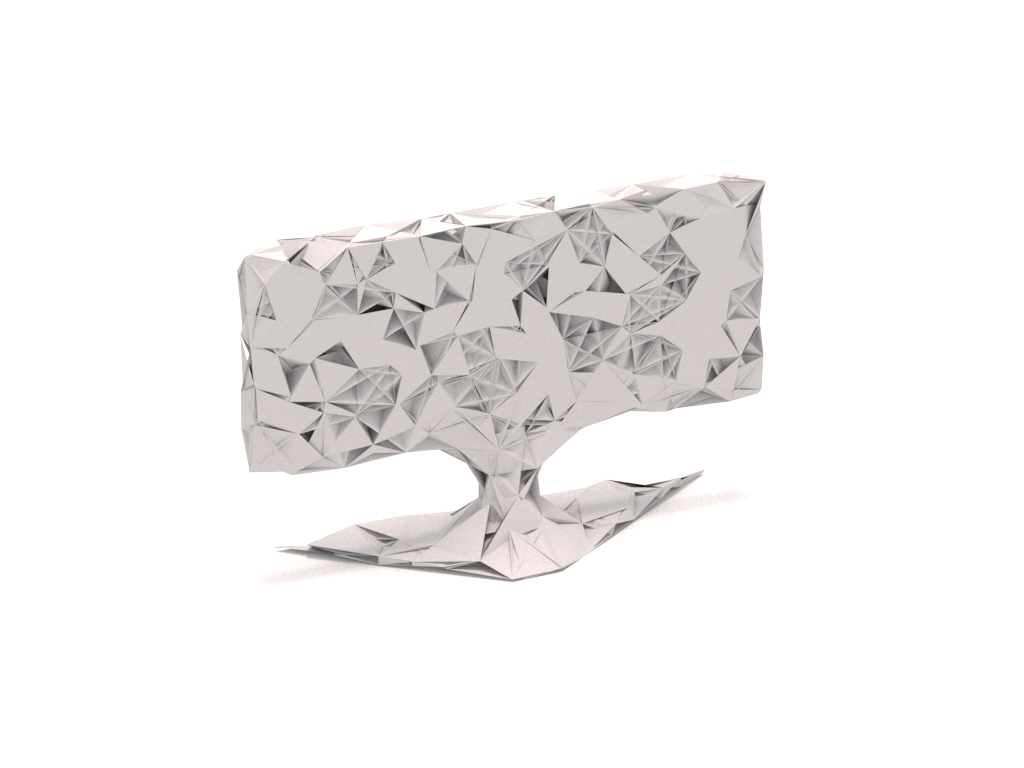}}
\vspace{-.1cm}
\vspace{-.1cm}
\centerline{\includegraphics[width=1.2\textwidth]{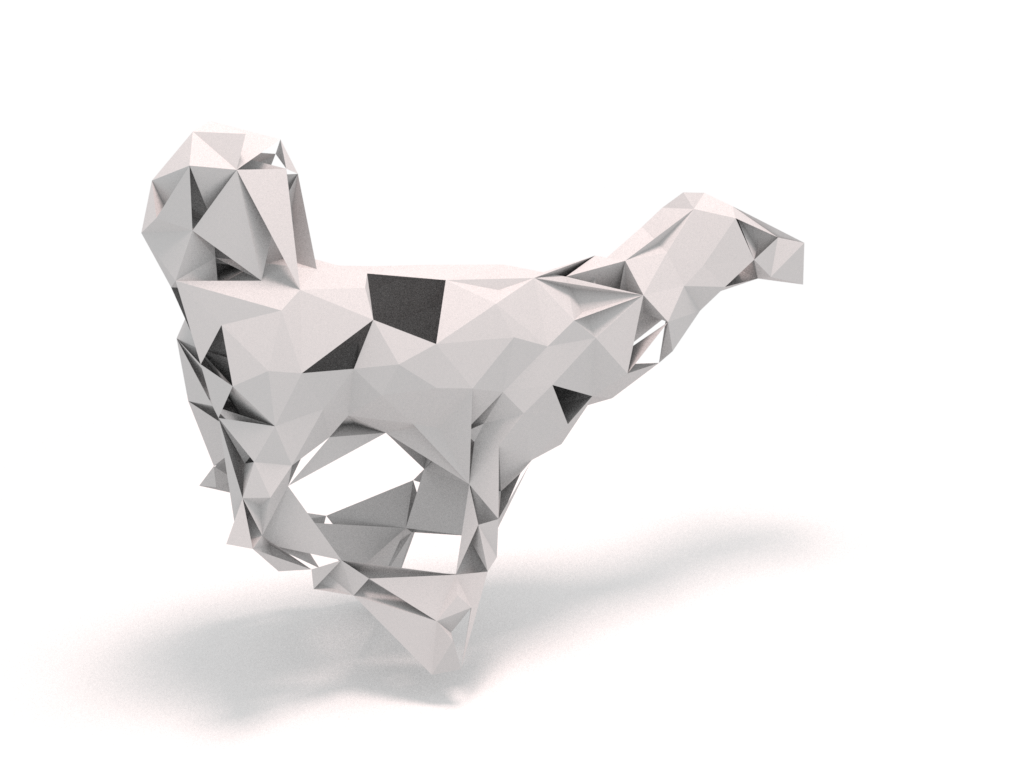}}
\centerline{\scriptsize {\makecell[c]{IER \\ \\}}}
\end{minipage}
\begin{minipage}{0.115\linewidth}
\centerline{\includegraphics[width=1.2\textwidth]{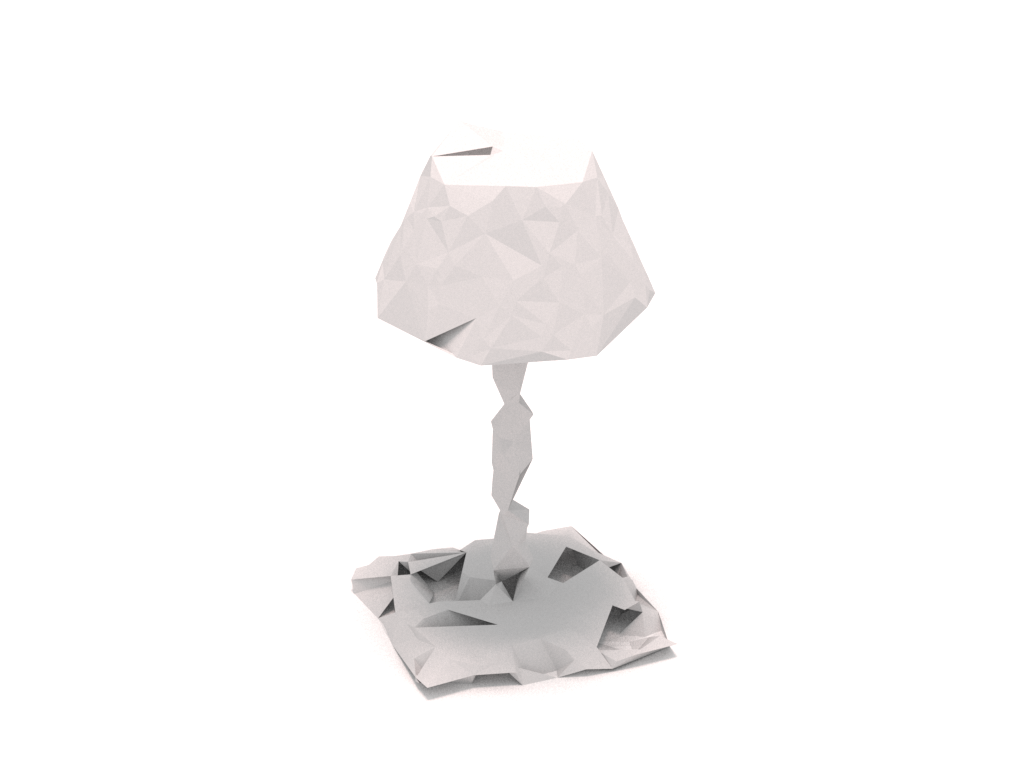}}
\vspace{-.1cm}
\centerline{\includegraphics[width=1.2\textwidth]{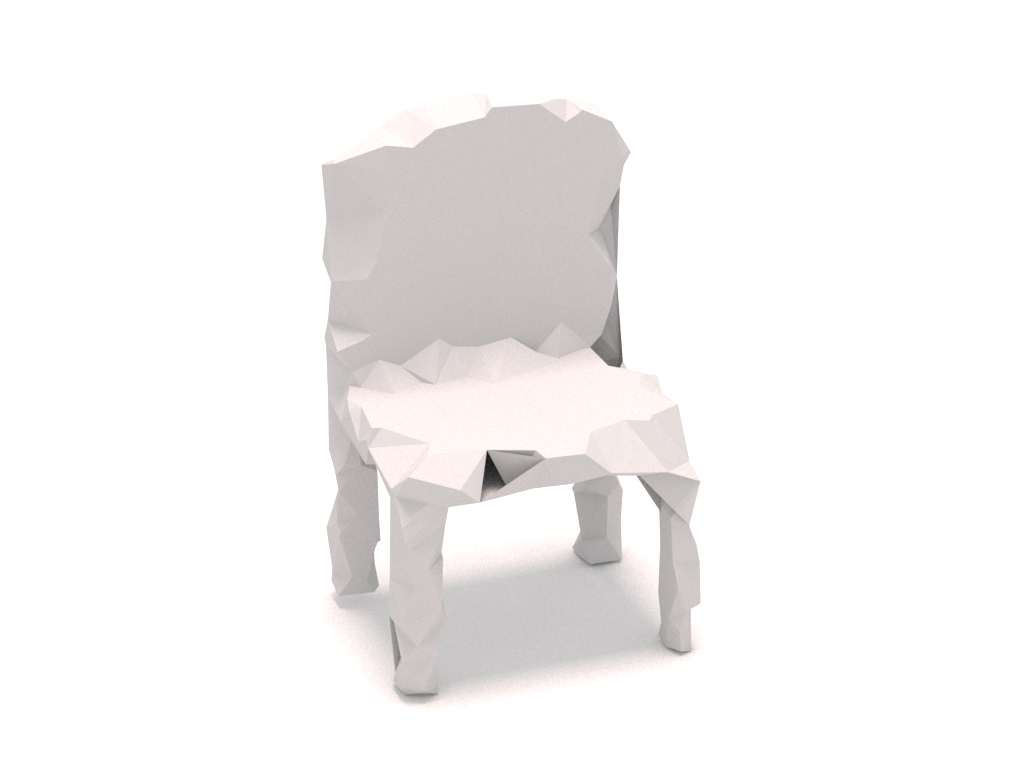}}
\vspace{-.1cm}
\centerline{\includegraphics[width=1.2\textwidth]{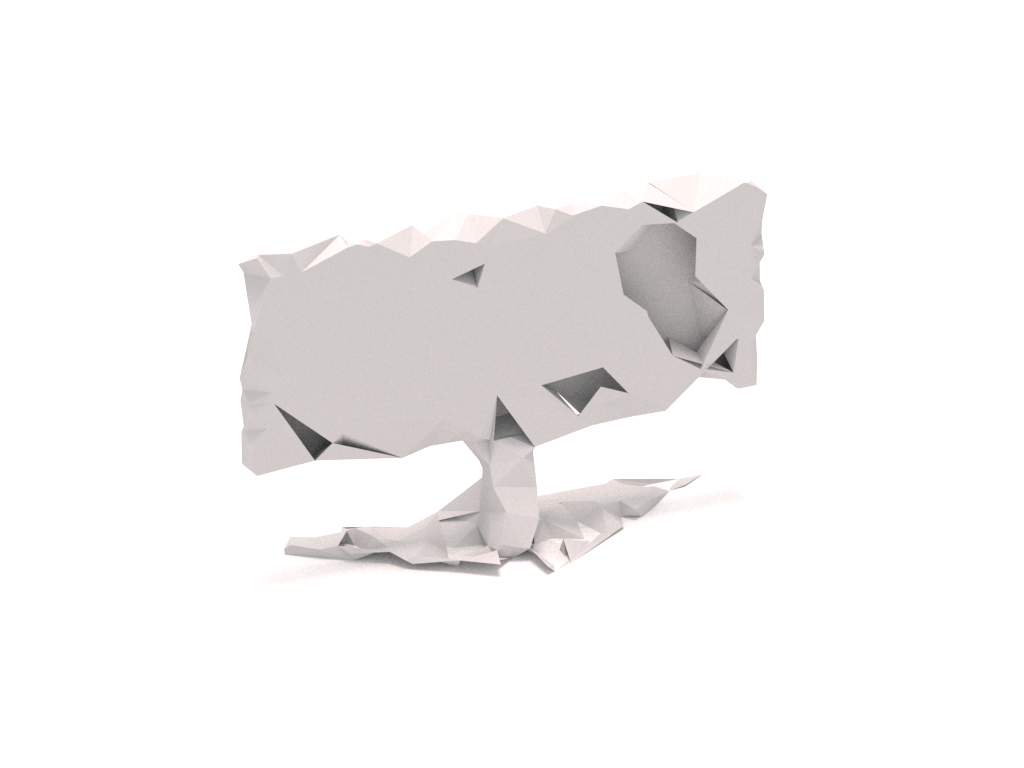}}
\vspace{-.1cm}
\vspace{-.1cm}
\centerline{\includegraphics[width=1.2\textwidth]{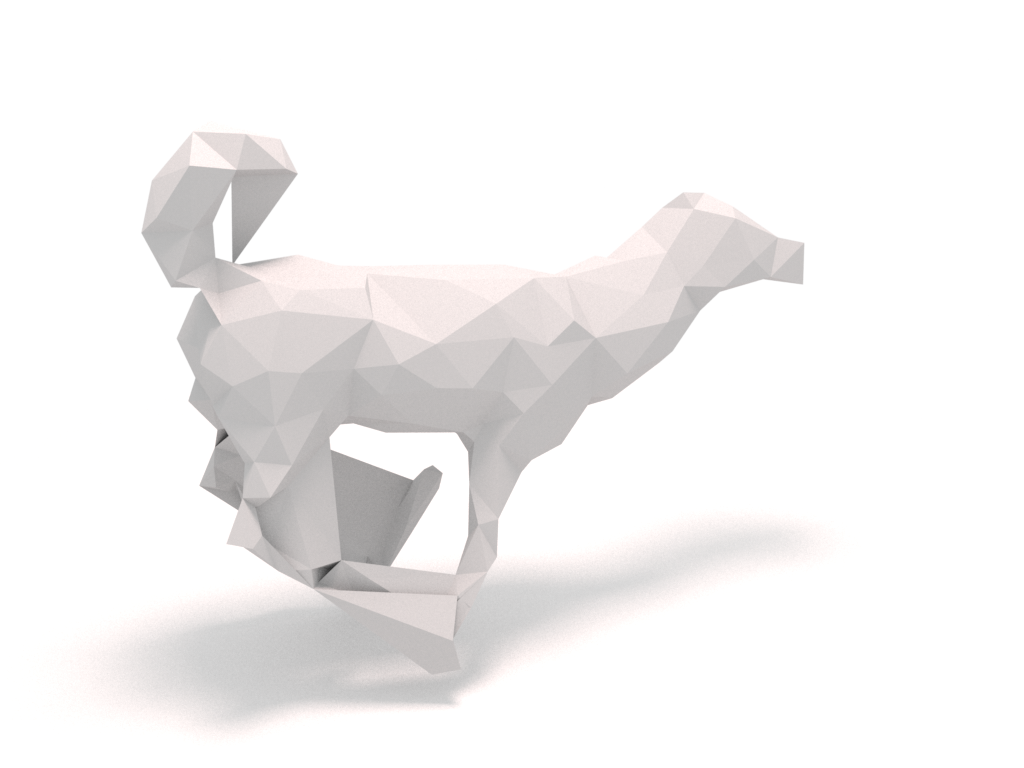}}
\centerline{\scriptsize{\makecell[c]{PointTriNet \\ \\}}}
\end{minipage}
\begin{minipage}{0.115\linewidth}
\centerline{\includegraphics[width=1.2\textwidth]{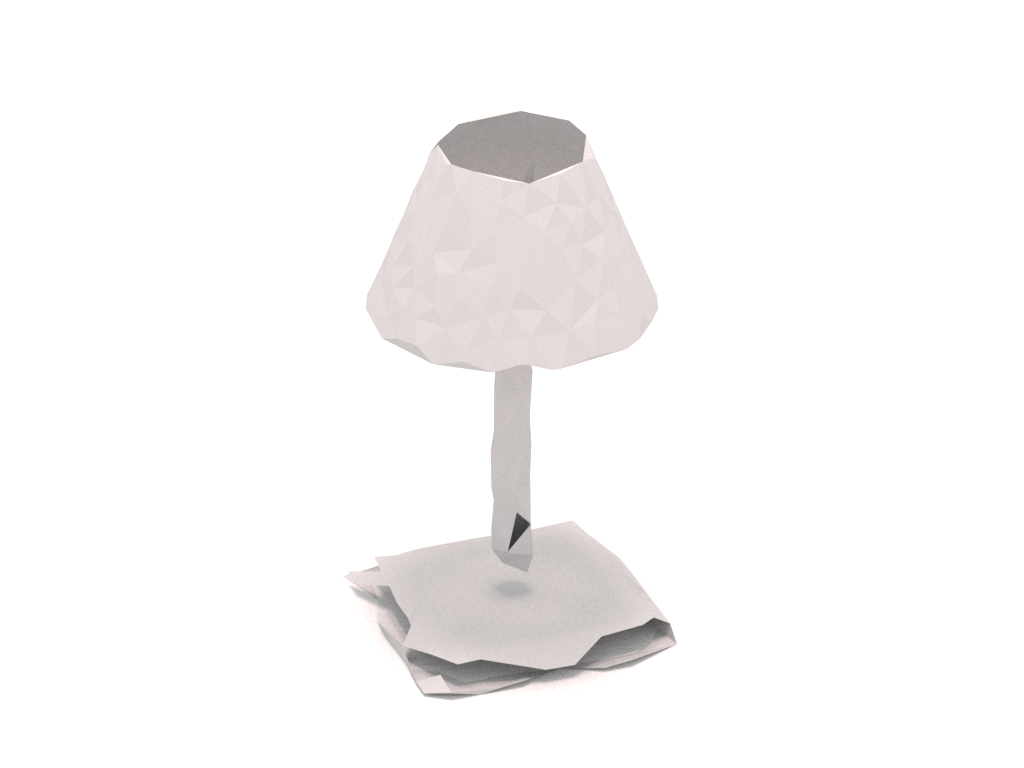}}
\vspace{-.1cm}
\centerline{\includegraphics[width=1.2\textwidth]{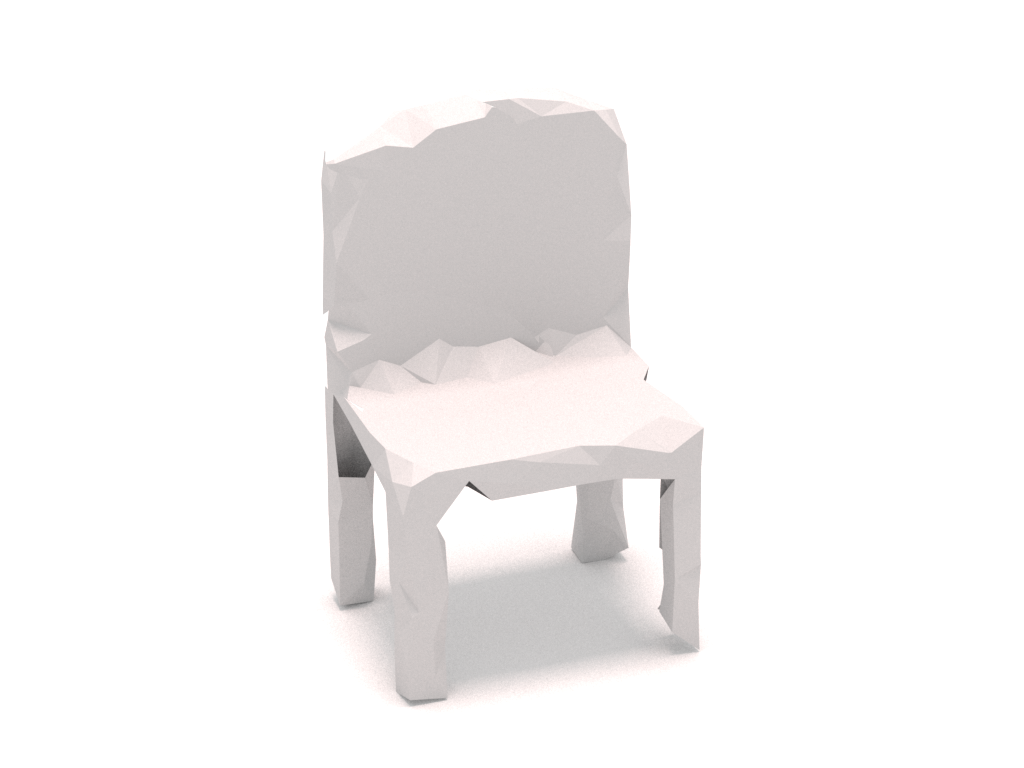}}
\vspace{-.1cm}
\centerline{\includegraphics[width=1.2\textwidth]{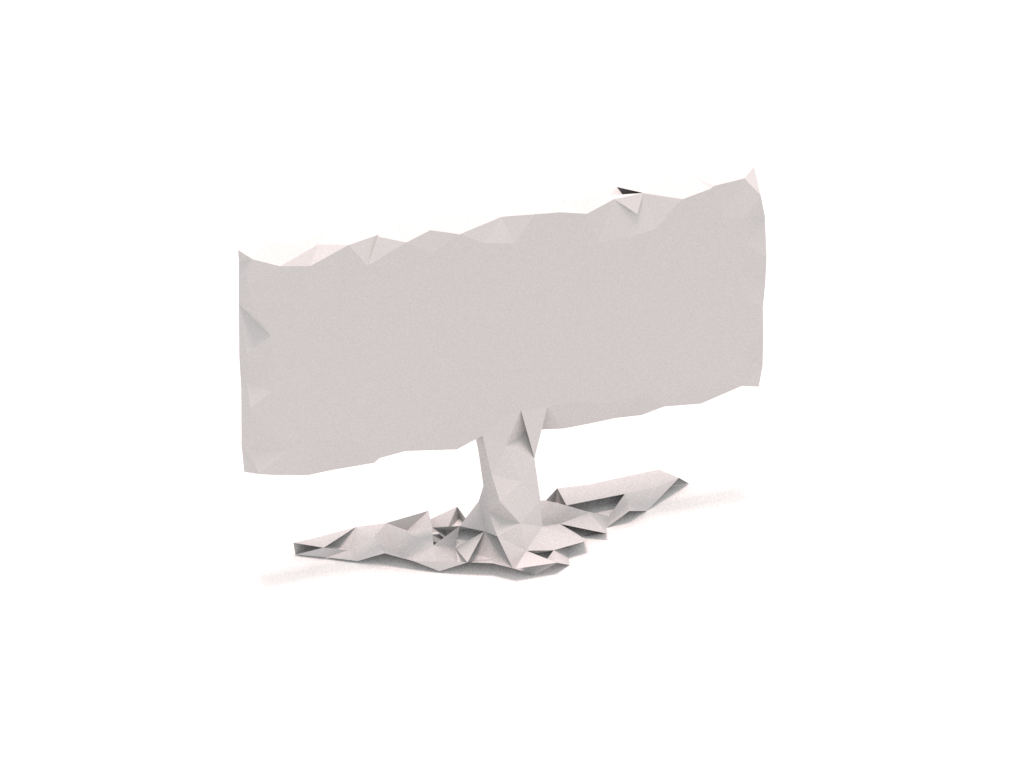}}
\vspace{-.1cm}
\vspace{-.1cm}
\centerline{\includegraphics[width=1.2\textwidth]{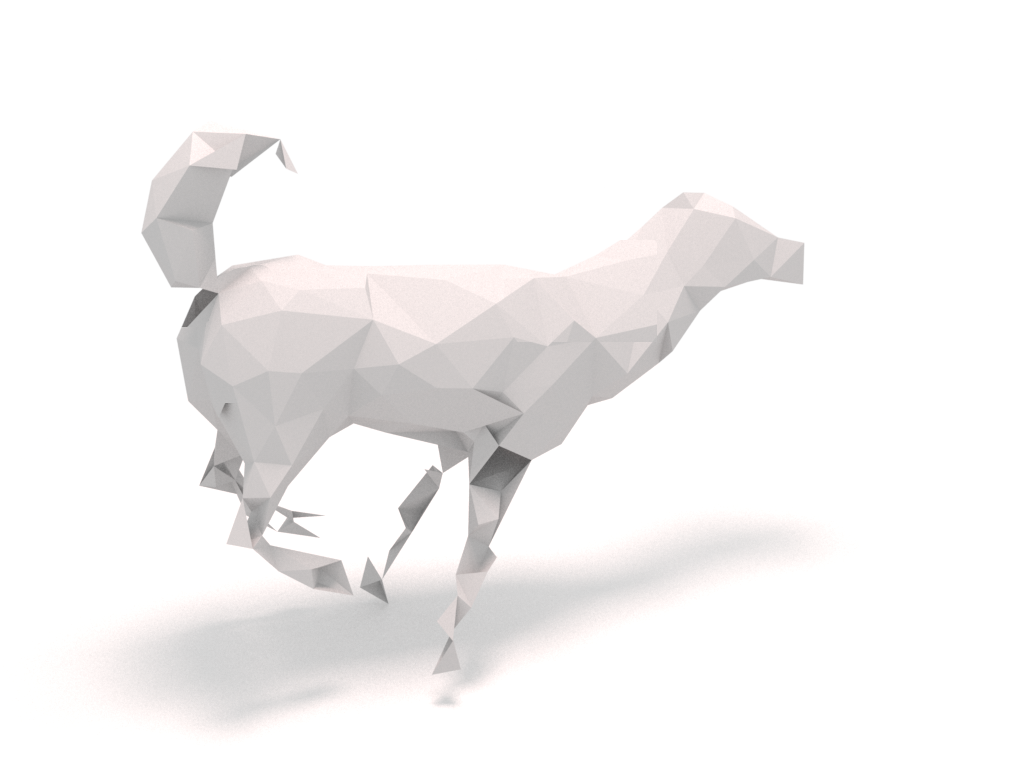}}
\centerline{\scriptsize{\makecell[c]{Ours \\ \\}}}
\end{minipage}
\begin{minipage}{0.115\linewidth}
\centerline{\includegraphics[width=1.2\textwidth]{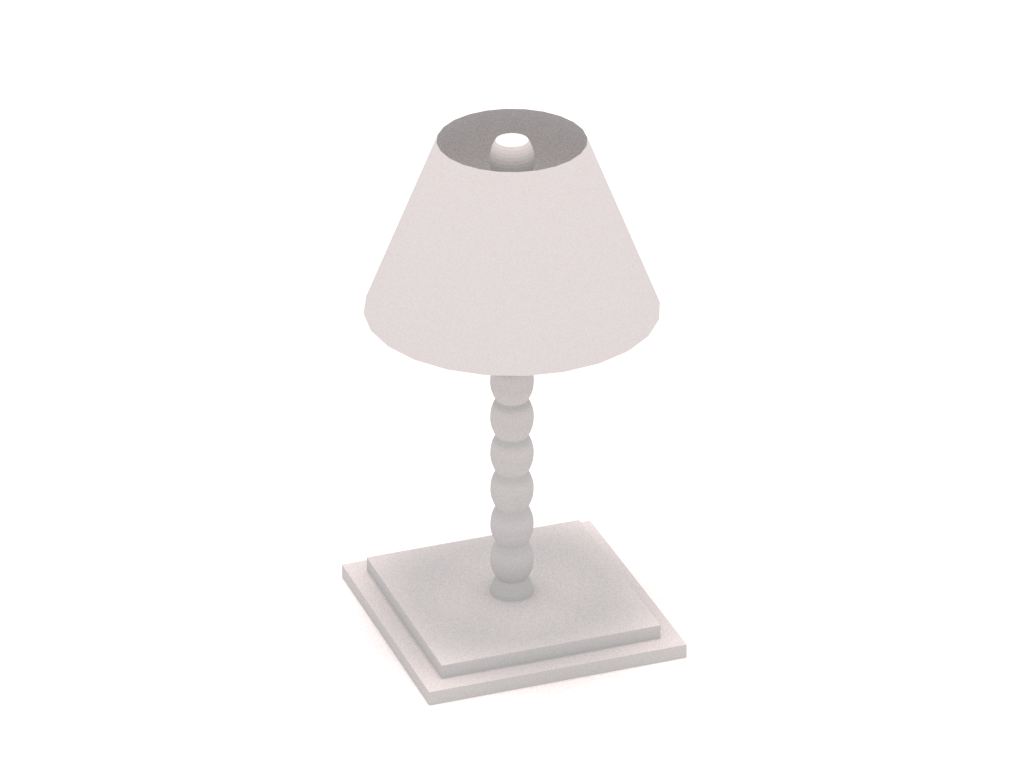}}
\vspace{-.1cm}
\centerline{\includegraphics[width=1.2\textwidth]{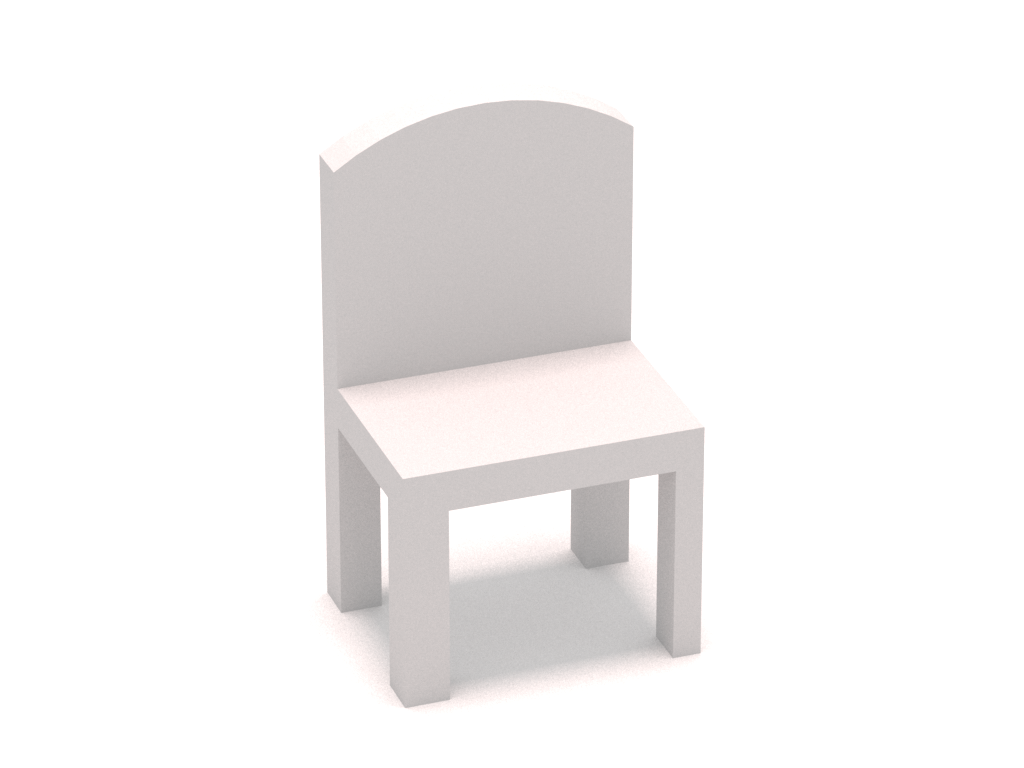}}
\vspace{-.1cm}
\centerline{\includegraphics[width=1.2\textwidth]{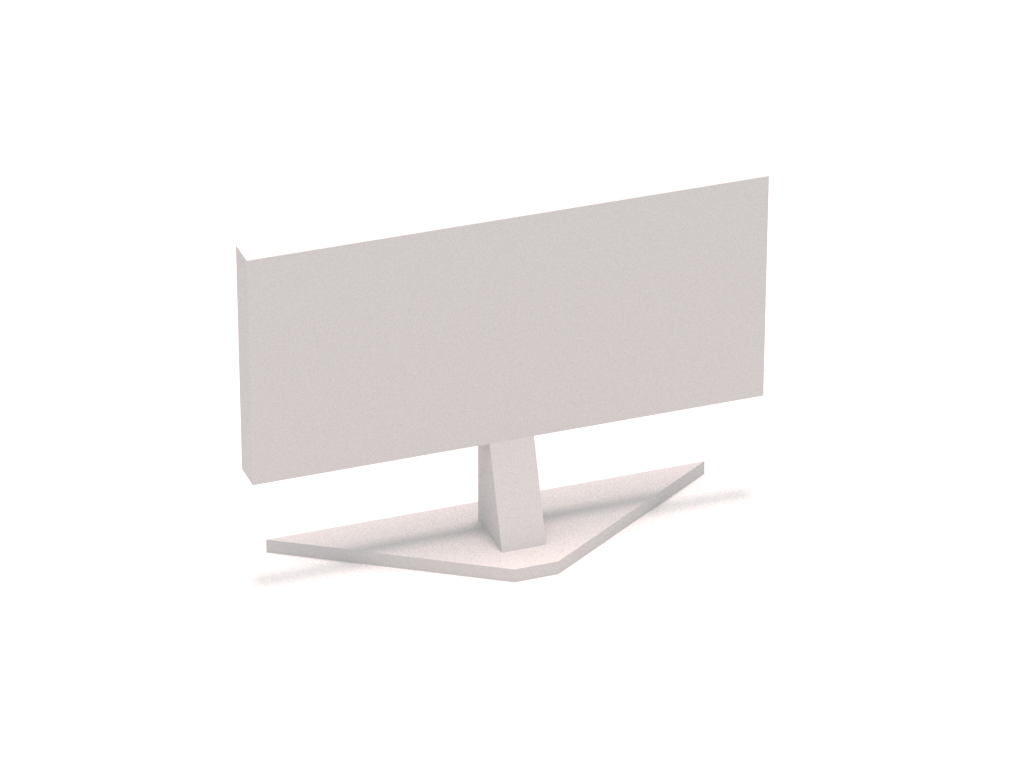}}
\vspace{-.1cm}
\vspace{-.1cm}
\centerline{\includegraphics[width=1.2\textwidth]{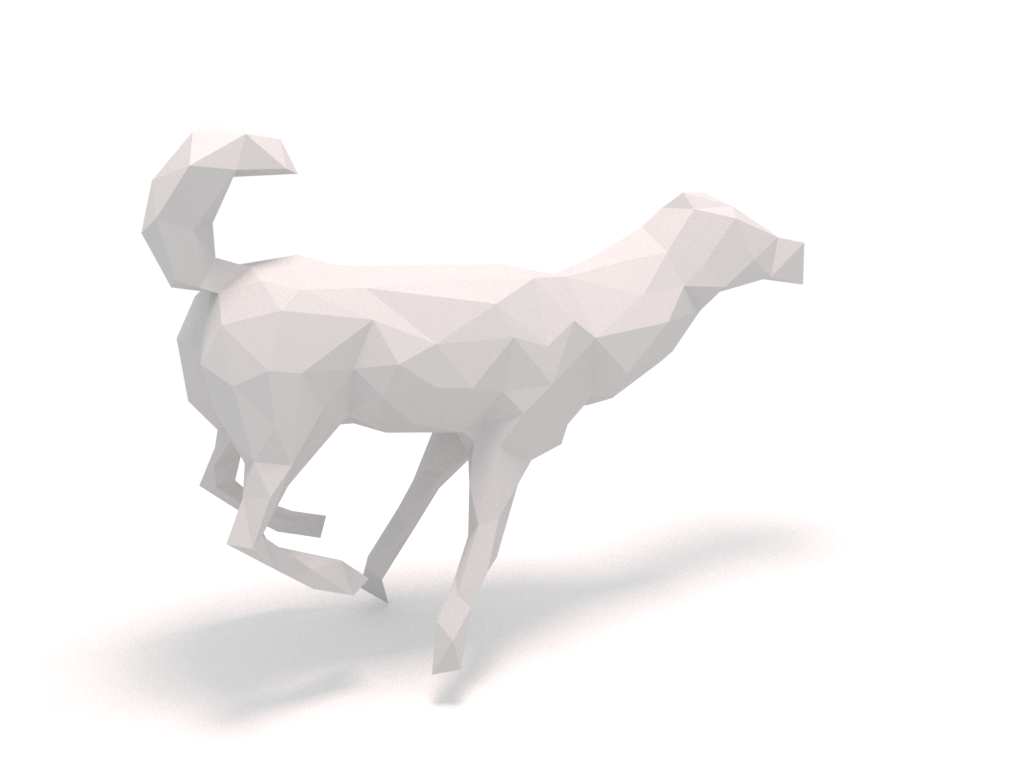}}
\centerline{\scriptsize {\makecell[c]{GT \\ \\}}}
\end{minipage}
\caption{Comparison of reconstruction visualizations with baselines.
Baselines are implicit reconstruction methods, including PSR, On-Surface, as well as explicit reconstruction methods such as BPA, IER and PointTriNet.
}
\label{fig:diff_vis}
\end{figure*}

\section{Experiments}
\subsection{Setup}
\textbf{{Datasets.}} To validate the effectiveness of the proposed method, we evaluate MergeNet for mesh reconstruction on SHREC dataset~\cite{lian2011shape} and ShapeNet dataset~\cite{chang2015shapenet}. SHREC dataset~\cite{lian2011shape} contains 30 classes of CAD models, with 20 objects in each class. The data split for train and test is consistent with the work~\cite{ezuz2017gwcnn}. For ShapeNet dataset~\cite{chang2015shapenet}, we take a subset of ShapeNet~\cite{chang2015shapenet} with the same train and test splitting as in IER~\cite{liu2020meshing}. 
We directly utilize the vertices of the meshes as the input sparse point cloud for SHREC dataset~\cite{lian2011shape},
while for the ShapeNet dataset~\cite{chang2015shapenet}, we sample 500 points on the model. 

\textbf{{Baselines.}} For the comparison with existing explicit reconstruction methods, we take rule-based BPA~\cite{bernardini1999ball} and learning-based PointTriNet~\cite{sharp2020pointtrinet} and IER~\cite{liu2020meshing} as baselines. 
Moreover, we also list traditional implicit PSR~\cite{kazhdan2013screened}
 and learning-based implicit reconstruction methods, such as NeuralTPS~\cite{chen2023unsupervised} and On-Surface~\cite{BaoruiMa2023ReconstructingSF}. We also provide On-Surface as a reference for comparison. 

\textbf{{Metrics.}} We adopt Chamfer Distance (CD) as a measurement of surface reconstruction accuracy.
The definitions of  L1 and L2 CD are the same as that in IER~\cite{liu2020meshing}. 
Additionally, we employ the F-score and Normal Consistency~(NC) metrics in the same way as in Ref~\cite{zimny2022points2nerf}. For the calculation of these two metrics, 10K points from both the reconstructed and GT mesh are sampled. The threshold for the matching judgment of F-score is set to $0.001$ in this work.

\subsection{Implementation details}  

Experiments are performed on an RTX 3090 GPU.
For parameter setting, $K$ of CEGM and $n$ of EEM are set to 32 and empirically 50. The threshold for edge filtering $d_{th}$ is set to 0.014. Adam is adopted as the optimizer. The learning rate is set to 0.00001 
with a decay of 0.3.

\begin{table}[h]
\renewcommand{\arraystretch}{1.5}
\caption{Comparison of runtime and performance.}
\begin{center}
\resizebox{0.85\linewidth}{!}{
\begin{tabular}{c|cccc|c}
\hline
\multirow{2}{*}{} & \multicolumn{4}{c|}{OnSurface (\#epoch v.s. time)}                                                                                       & \multirow{2}{*}{Ours} \\ \cline{2-5}
                  & \multicolumn{1}{c}{500 } & \multicolumn{1}{c}{1k} & \multicolumn{1}{c}{3k} & 30k (default) &                       \\ \hline
Times(s)          & \multicolumn{1}{c}{49.4}       & \multicolumn{1}{c}{56.9}        & \multicolumn{1}{c}{145.1}       & 345.7        & 27.3                  \\ 
\hline
\end{tabular}
}
\label{Fig:diff_epoch}
\end{center}
\end{table}

\subsection{Quantitative results}

We report the quantitative comparison results between our method and baselines on ShapeNet and SHREC in Tab.\ref{L1_ShapeNet} and Tab.\ref{L2_SHREC}. We can find that our MergeNet achieved the best performance in almost all metrics compared with explicit baselines.
It also shows competitive performance compared with implicit methods, which often require more computations to evaluate the function at each point in space, which can be computationally intensive and slow down the reconstruction process.
To demonstrate this, we compared the processing time with On-Surface~\cite{BaoruiMa2023ReconstructingSF} in Tab.~\ref{Fig:diff_epoch}. In Fig.~\ref{fig:diff_vis}, we presented the qualitative results of On-Surface after 500 epochs. It can be observed that On-Surface exhibits the issue of information loss. 
On-Surface requires multiple epochs to generate a high-quality SDF and then uses marching cubes to generate the surface. Both of the two steps of  On-Surface are highly time-consuming. Our approach requires significantly fewer computational resources and less time than On-Surface~\cite{BaoruiMa2023ReconstructingSF}.

\subsection{Qualitative results} 
\begin{figure}[htbp]
  \centering
  \includegraphics[width=1\linewidth]{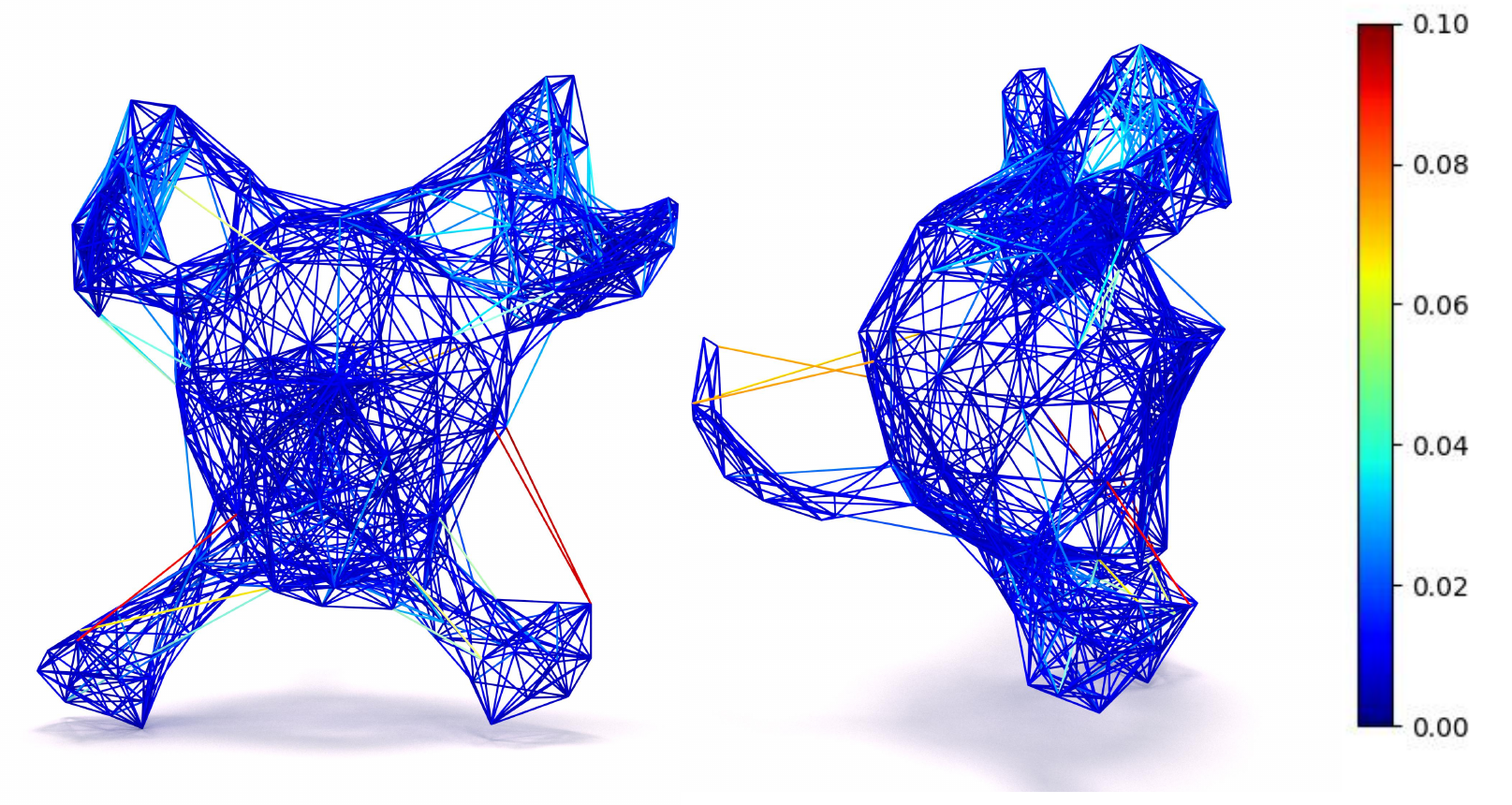}
  \caption{Error map between the predicted edge-to-face distance and GT. Colder color indicates smaller error.}
  \label{fig:edges}
\end{figure}

In Fig.~\ref{fig:diff_vis}, we demonstrate qualitative results for the ShapeNet ~\cite{chang2015shapenet} and SHREC~\cite{lian2011shape} datasets. When the point cloud is sparse, the traditional Poisson reconstruction algorithm struggles with sparse point clouds in a non-adaptive manner and results in discontinuous flakes on the surface.
In contrast, learning-based methods are more localized. The performance of the IER method in reconstructing surfaces is significantly worse when the point clouds are sparse.
PointTriNet generates initial triangles greedily and then iteratively generates triangles, which may result in the generation of large gaps. If filled, PointTriNet's hole-filling logic leads to apparently incorrect triangles. In contrast, our approach predicts lines directly and uses modern data-driven learning techniques to generate lines end-to-end, followed by post-processing to generate geometric surfaces that respect the original surface. 
We have provided additional results on real LiDAR data in the supplementary materials.  Please refer to them.
We also visualize the edge-wise error in Fig.~\ref{fig:edges}.

\begin{figure}[h]
  \centering
  \includegraphics[width=1\linewidth]{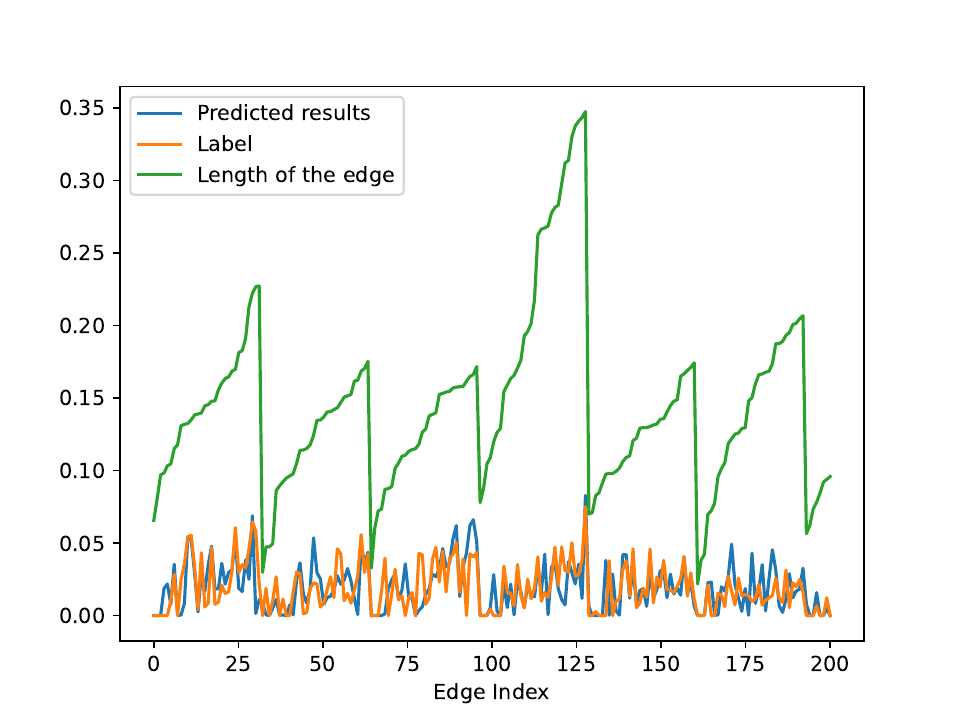}
  \caption{The relationship between edge length, prediction and the ground truth distance of edge-to-face for the first 200 candidate edges of a trained model. This indicates that the model learns to predict the  edge-to-face distance for edges with various lengths.}
  \label{fig:edgeLen_pre_label}
\end{figure}

Moreover, we select the first 200 candidate edges to show the relationship between label~(the ground truth distance of edge-to-face), prediction and edge length is shown in Fig.~\ref{fig:edgeLen_pre_label}. We adopt a KNN-based approach to select the candidate edges. Through visual inspection of the results shown in the figure, it is evident that our network  can predict well for candidate edges of varying lengths.

\begin{table}[htb]
\caption{Ablation of edge embedding module.}
\begin{center}
\begin{tabular}{c|cc}
\hline
    Edge embedding      & w/o & w/ \\
\hline
$\mathrm{L_2 CD}$($\times 10^{-4}$)  & 3.96 & 1.14\\
\hline
\end{tabular}
\label{fig:embedding}
\end{center}
\end{table}

\begin{table}[htb]
\caption{The performance of the proposed MergeNet on ShapeNet with varying sampled points.}
\begin{center}
\begin{tabular}{c|cccc}
\hline
 points & 250 & 500 & 1k & 2k \\
\hline
$\mathrm{L_2 CD}$($\times 10^{-4}$)  & 1.18 & 1.14 & 1.07  & 0.91 \\
\hline
\end{tabular}
\label{tab:diff_points}
\end{center}
\end{table}
\begin{figure}[htb]
\begin{minipage}{0.24\linewidth}
\centerline{250 points}
\centerline{\includegraphics[width=\textwidth]{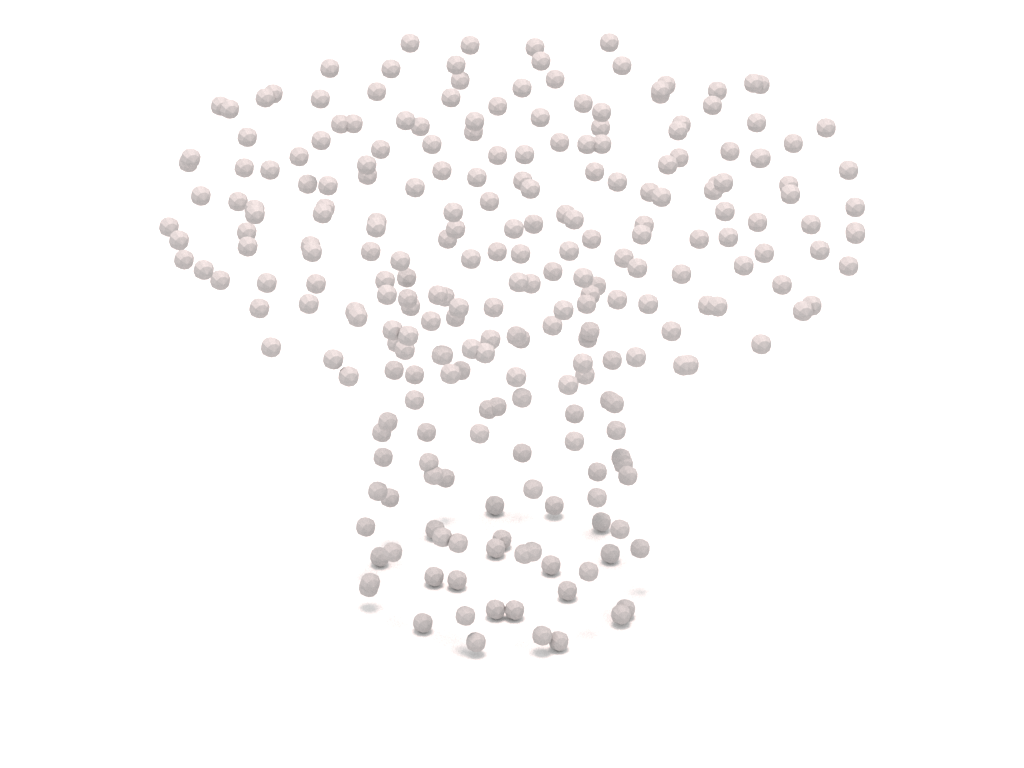}}
\centerline{\includegraphics[width=\textwidth]{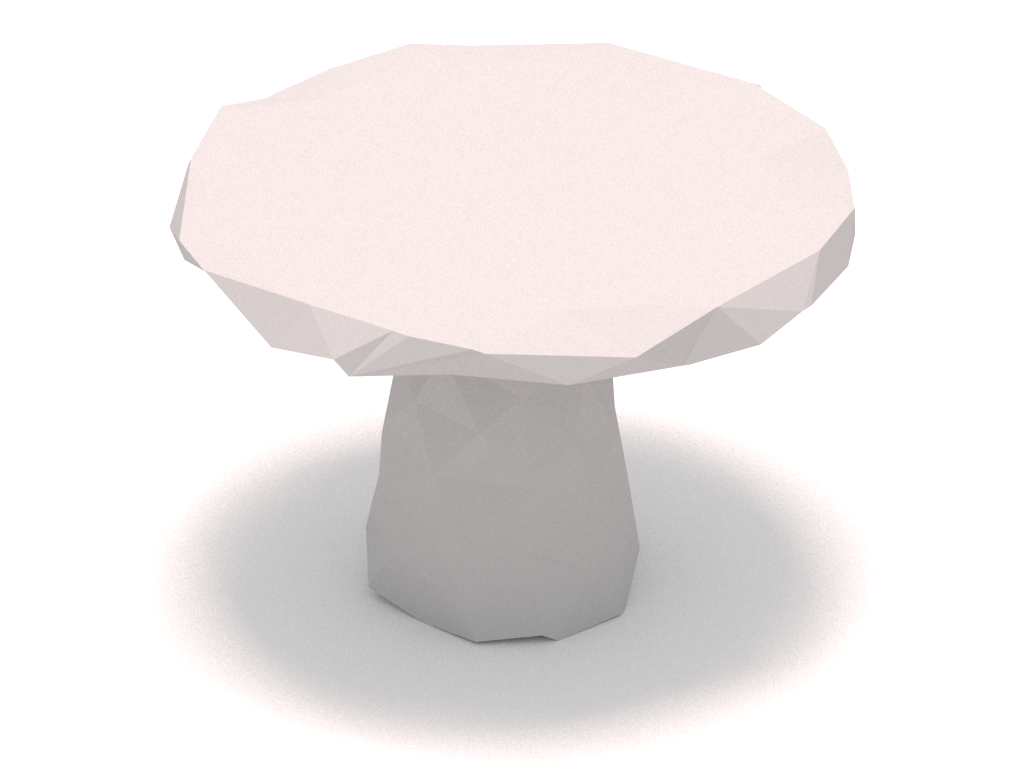}}
\end{minipage}
\begin{minipage}{0.24\linewidth}
\centerline{500 points}
\centerline{\includegraphics[width=\textwidth]{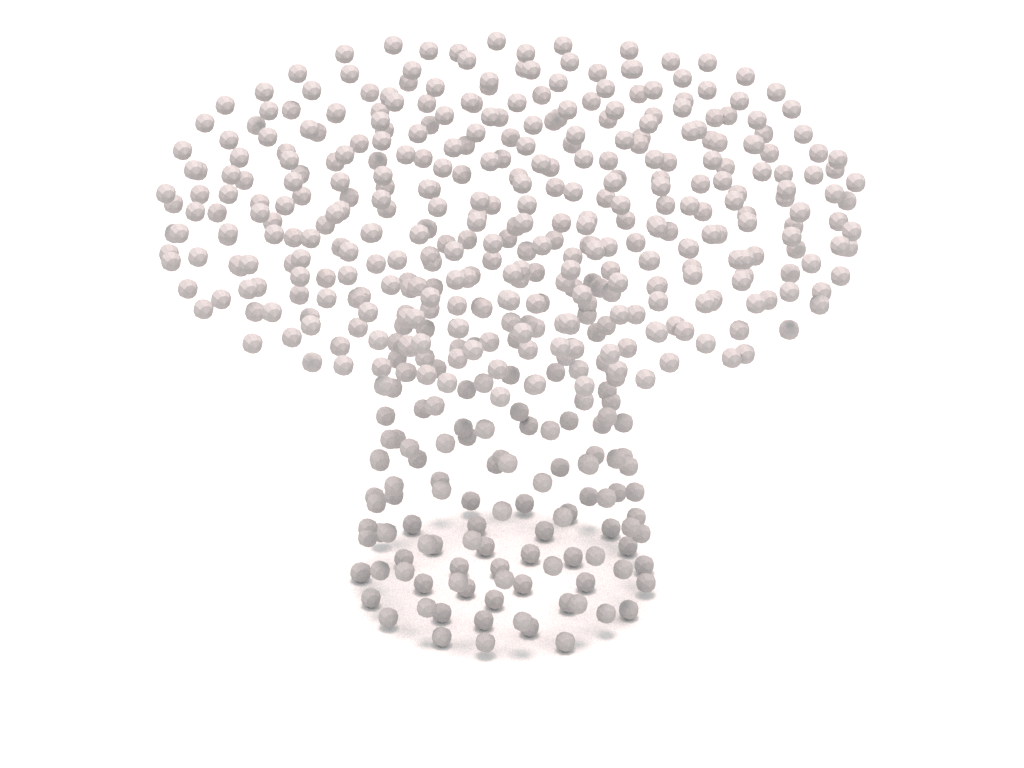}}
\centerline{\includegraphics[width=\textwidth]{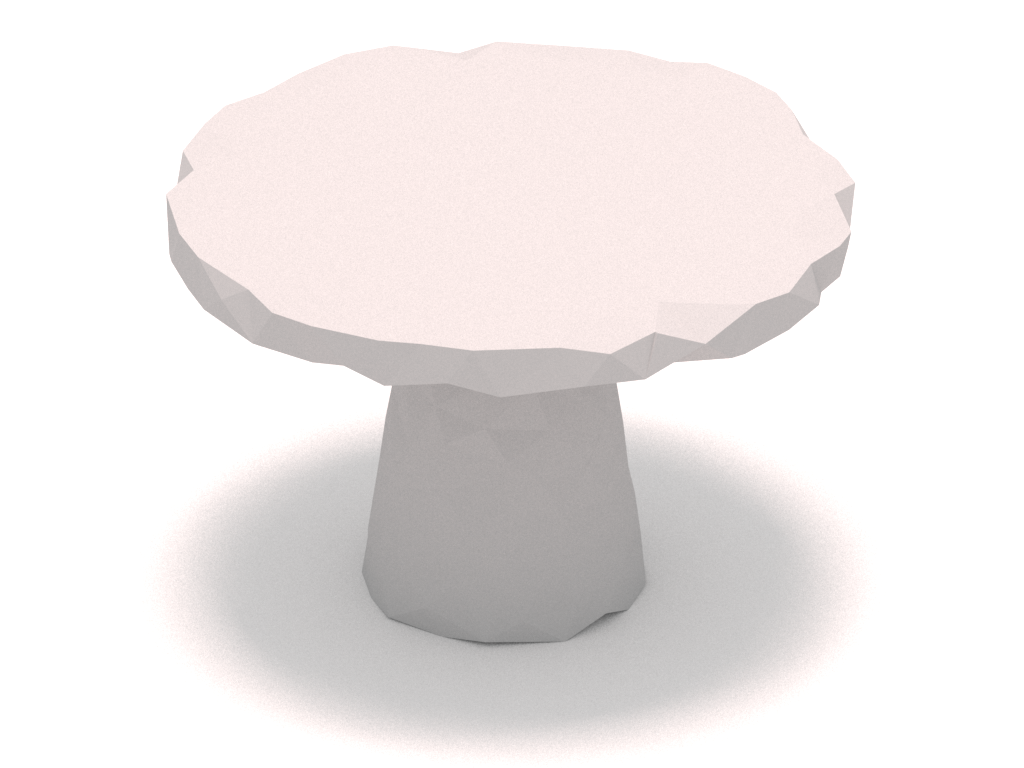}}
\end{minipage}
\begin{minipage}{0.24\linewidth}
\centerline{1000 points}
\centerline{\includegraphics[width=\textwidth]{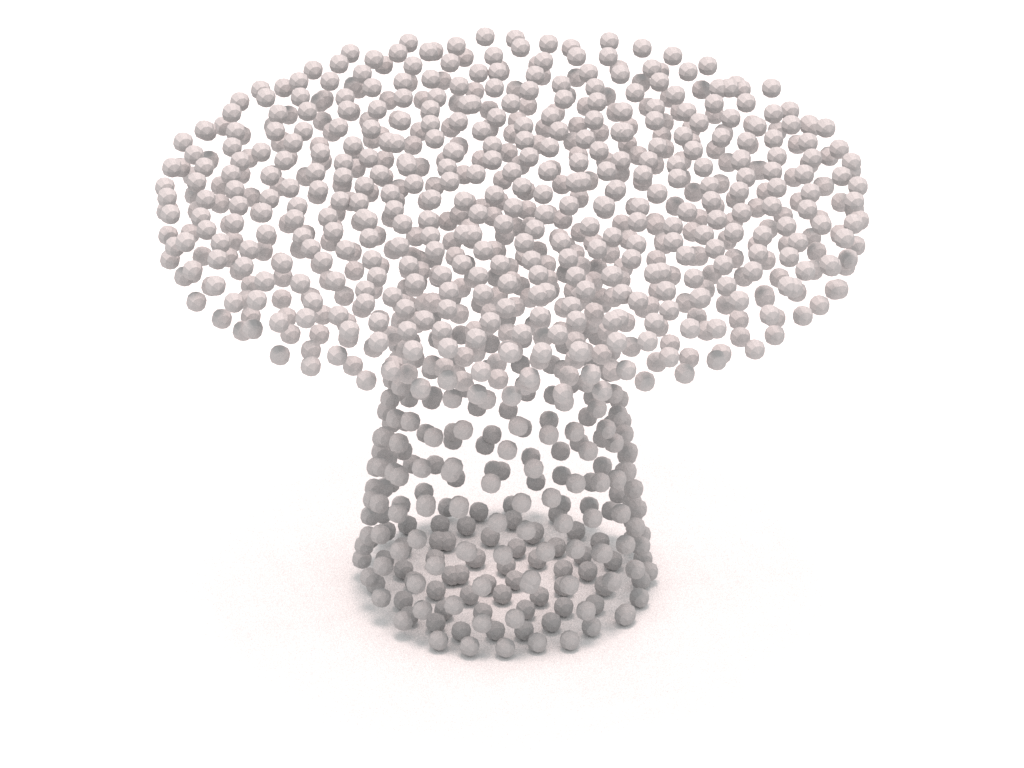}}
\centerline{\includegraphics[width=\textwidth]{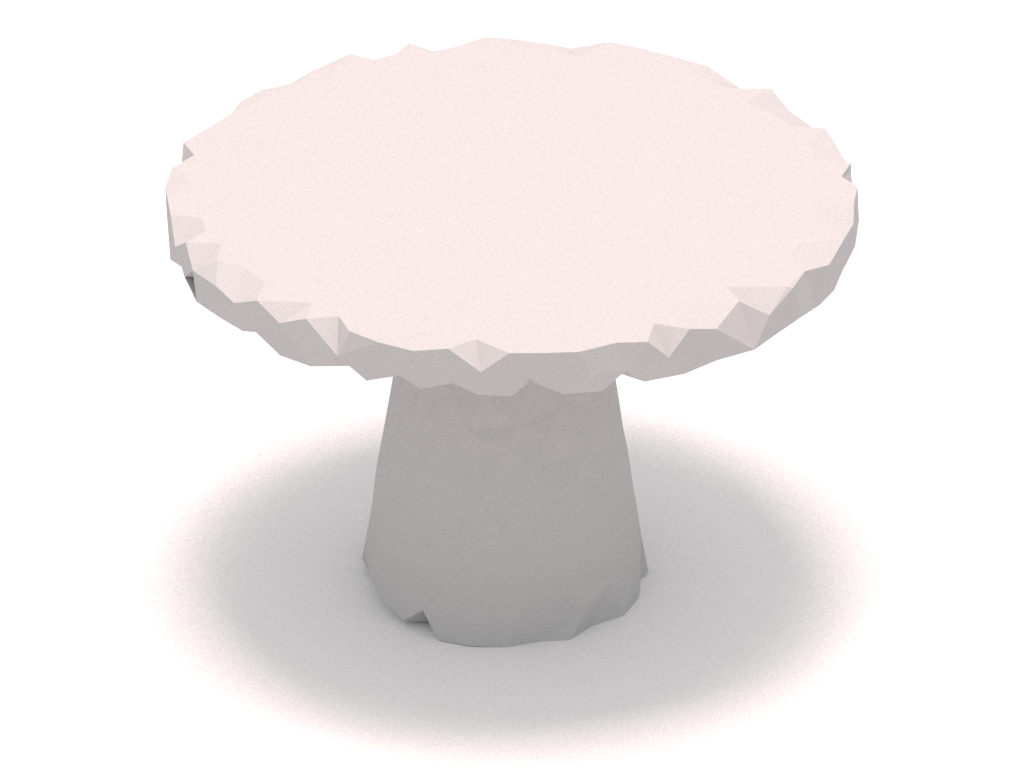}}
\end{minipage}
\begin{minipage}{0.24\linewidth}
\centerline{GT}
\centerline{\includegraphics[width=\textwidth]{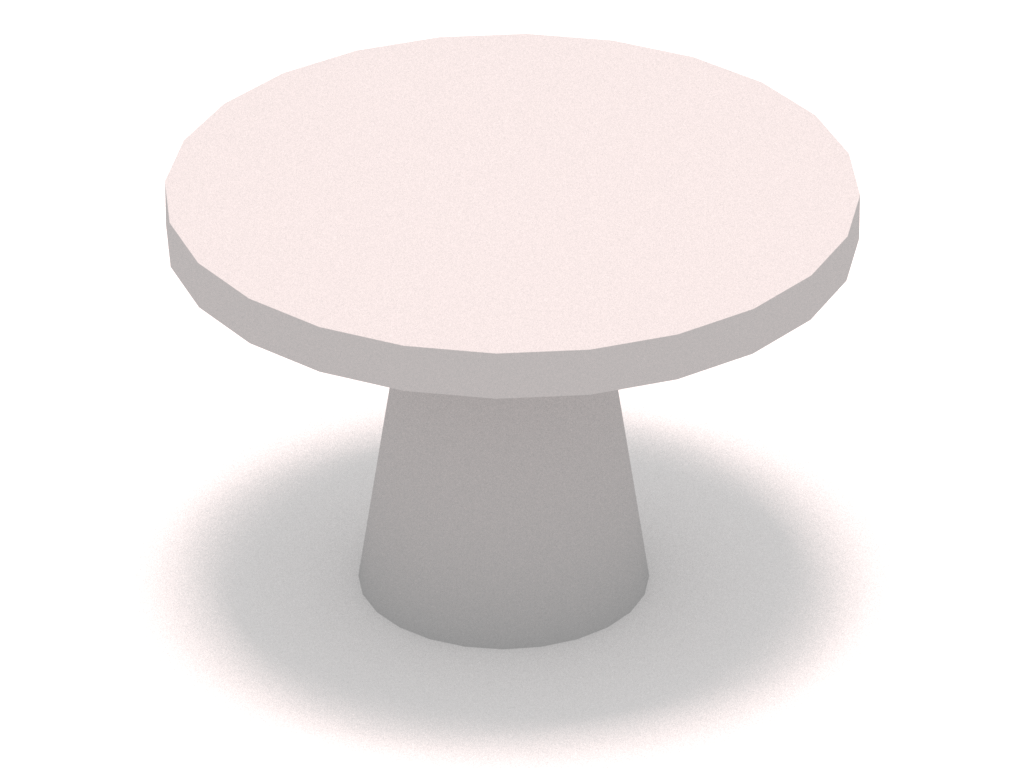}}
\end{minipage}
\caption{Reconstruction results with different numbers of points.}
\label{fig:diff_points}
\end{figure}

\subsection{Ablation study}
We conduct experiments to verify the reconstruction performance under different levels of point cloud sparsity and to evaluate the effectiveness of our edge embedding strategy. 
The results are reported in Tab.~\ref{fig:embedding}. The $\mathrm{L_2}$ CD degrades greatly without the edge embedding, indicating that the canonical normalization for edge embedding endows the networks with the ability to better embed local information.

To further verify the generalization ability of the network, we evaluate the performance of our model trained with various numbers of point clouds for reconstruction tasks. Specifically, we train and test the model on 250, 500, 1000, and 2000 points respectively and present the quantitative results in Tab.~\ref{tab:diff_points} and the visualization results in Fig.~\ref{fig:diff_points}. We can find that Merge performs better as the number of points increases while keeping robust for sparse input.

\section{Conclusion}
In this paper, we introduce MergeNet, a novel approach for reconstructing meshes from point clouds. 
MergeNet tackles the challenges of point cloud sparsity and the efficiency of isosurface extraction by addressing mesh reconstruction as the edge connectivity prediction problem. Extensive experiments and qualitative visualizations demonstrate the superiority of the proposed edge-based approach in both quality and efficiency of generated meshes, especially for sparse point clouds. 
Ablation studies on the edge embedding factor verified the effectiveness of canonical normalization. 
Although superior reconstruction performance to SoTA methods is demonstrated, some holes still exist which is a common problem for all explicit approaches.
We will incorporate more effective hole-filling methods for better performance.
Furthermore, given MergeNet's capacity to predict point connectivity, we plan to explore this potential to benefit other point cloud understanding tasks in various fields.

\section*{Acknowledgment}
This research was supported by the Natural Science Foundation of China under Grant No. 62306059, 61936002, 62102061 and No. T2225012; the National Key R$\&$D Program of China under Grant No. 2021YFA1003003.

\bibliographystyle{ieeetr} 

\bibliography{icme2023template}


\end{document}